# Quantum-classical physics-informed Kolmogorov-Arnold networks for PDEs

**Xiang Rao[1,2,3,4,5]*, Yuxuan Shen[5]**

1. School of Petroleum Engineering, Yangtze University, Wuhan 430100, China

2. School of Computer Science, Yangtze University, Jingzhou 434023, China

3. State Key Laboratory of Low Carbon Catalysis and Carbon Dioxide Utilization (Yangtze University), Wuhan 430100, China

4. Western Research Institute, Yangtze University, Karamay 834000, China

5. College of Future Technology, Yangtze University, Wuhan 430100, China

*Corresponding author: Xiang Rao (raoxiang0103@163.com, raoxiang@yangtzeu.edu.cn)

Abstract: We develop QCPIKAN, the first quantum-classical physics-informed Kolmogorov-Arnold network designed to solve partial differential equations (PDEs). Built upon Chebyshev-polynomial KAN layers and parameterized quantum circuits, this hybrid framework embeds physical constraints into the training loss to enforce physical consistency. Our theoretical investigations grounded in approximation theory prove that this design accelerates high-frequency error convergence to an exponential rate and effectively mitigates numerical dispersion. We validate the framework across three typical seepage scenarios in porous media, including single-phase flow, component transport and two-phase flow. Compared with existing quantum-classical physics-informed neural networks, QCPIKAN achieves superior performance in global prediction accuracy, local error control, dynamic evolution tracking and displacement front localization. This work provides a robust and efficient alternative for solving complex PDEs.

## Introduction

In recent years, physics-informed neural networks (PINNs) have attracted broad attention in scientific machine learning because they can directly embed physical constraints, including partial differential equations (PDEs), boundary conditions and initial conditions, into the neural-network training process. By introducing the physical residuals associated with governing equations into the loss function, PINNs allow the network to fit observational data while satisfying physical conservation laws, thereby enabling numerical solution and parameter inversion for complex physical fields [1]. Compared with purely data-driven models, PINNs make explicit use of prior physical information and therefore show stronger generalization and physical consistency when training data are limited or observations are sparse [2].

Owing to this physics-constrained mechanism, PINNs have gradually been applied to complex scientific-computing problems in fluid mechanics, bioengineering, manufacturing systems and related fields. For example, in mechanical fault diagnosis, PINNs can accurately identify time-varying mesh stiffness even when the training data contain only a small number of meshing cycles and are contaminated by random noise; the maximum prediction error is controlled within 4.7%, demonstrating the ability of physical constraints to resist noise disturbance [3]. In parameter inversion for complex biological dynamical systems, multiple key dynamical parameters can be recovered accurately from only a limited number of discrete observation points, indicating the strong parameter-identification ability of PINNs in small-sample nonlinear inverse problems [4]. For manufacturing-system state prediction, Hua et al. [5] proposed an uncertainty-evaluation-based weighted-loss strategy to address inaccurate physical models and noisy data; by quantifying prediction-error variance and dynamically adjusting the weights of data and physics losses, the method improves both accuracy and stability. In inverse problems for frictional-contact temperature fields, Xia and Meng [6] constructed a hybrid loss function that combines observational data and physical residuals, enabling stable temperature-field prediction when all input parameters are unknown; the average prediction error was reduced from 5.16% for a conventional data-driven model to 0.79%, substantially improving accuracy under complex boundary conditions. PINNs have also shown promising performance in subsurface-flow simulation and permeability-parameter inversion. Tartakovsky et al. [7] showed that PINNs can recover underground flow states from sparse observations and jointly infer permeability fields and nonlinear constitutive relations. Hanna et al. [8] used residual-based adaptive sampling to dynamically add collocation points in high-residual regions and improve front capture. Fuks and Tchelepi [9] introduced artificial viscosity to improve PINN accuracy for the Buckley-Leverett (BL) equation in two-phase porous-media flow. Xu et al. [10] improved PINN accuracy by treating the weak form of the BL equation rather than the original differential equation. Almajid and Abu-Al-Saud [11] applied PINNs to both forward and inverse BL problems, and Lehmann et al. [12] investigated a mixed pressure-velocity PINN formulation for heterogeneous single-phase reservoir flow. Inspired by convolutional neural networks and the finite-volume method, Fang [13] proposed an efficient hybrid PINN with a guaranteed convergence rate by replacing automatic differentiation with approximate differential operators. Similarly, Yan et al. [14] and Lin et al. [15] further improved computational performance under complex heterogeneous-reservoir or discrete-fracture conditions by introducing finite-volume discretization, dynamic loss weighting and hierarchical physical constraints.

Although PINNs exhibit substantial potential for complex seepage problems, their function-representation capability still depends largely on the conventional multilayer perceptron (MLP) architecture. When dealing with strongly nonlinear, multiscale and complex-interface problems, traditional MLPs are prone to spectral bias, difficult gradient propagation and parameter redundancy, which limits the training stability and generalization ability of PINNs in complex physical fields.

To overcome the limitations of traditional MLPs in function representation, parameter efficiency and interpretability, Kolmogorov-Arnold networks (KANs), constructed on the basis of the Kolmogorov-Arnold representation theorem, provide a new representation paradigm by replacing fixed activation functions and linear weight mappings with learnable one-dimensional nonlinear functions [16]. Motivated by this idea, physics-informed Kolmogorov-Arnold networks (PIKANs), in which KANs replace traditional MLPs, have received increasing attention. Compared with conventional PINNs, PIKANs show advantages in parameter efficiency, feature representation and noise robustness when solving forward and inverse PDE problems [17], [18]. For instance, Shuai and Li [19] achieved high-accuracy prediction of power-system dynamics using a smaller network scale; Jiang et al. [20] proposed a hybrid KAN-ANN model for short-term load forecasting that improves prediction accuracy while preserving interpretability; and Lin et al. [21] developed a KAN-based difference-feature enhancement algorithm with uncertainty guidance for semi-supervised remote-sensing change detection. Huang et al. [22] and Jiang et al. [23] coupled PIKANs with finite-element methods for deformation analysis of complex shell structures and further applied them to ship-hull structural engineering, verifying their predictive ability and parameter-expression advantages under complex boundary conditions. PIKANs have also been applied to landslide time-to-failure prediction and multi-material elasticity, where results show good stability and generalization for complex interfaces, strong nonlinearity and multi-material coupling [24], [25]. For subsurface flow, Rao et al. [26] combined PIKANs with a mixed pressure-velocity formulation for heterogeneous-reservoir simulation and found better computational performance than traditional PINNs. Imankulov et al. [27] conducted a sensitivity study of PIKANs for Buckley-Leverett shocks and showed that PIKANs can achieve accuracy comparable to PINNs with fewer parameters. Wang et al. [28] combined asymptotic analysis with PIKANs to obtain label-free solutions of single-phase seepage equations with source and sink terms.

In the current noisy intermediate-scale quantum (NISQ) era [29], [30], variational quantum algorithms (VQAs) have developed into an important hybrid-computing framework whose core idea is to iteratively update parameterized quantum circuits using classical optimizers [30]. Within this framework, many representative subalgorithms have emerged, including variational quantum eigensolvers (VQEs) [31], [32], quantum support vector machines (QSVMs) [33], [34], variational quantum linear solvers (VQLSs) [35]-[40], quantum approximate optimization algorithms (QAOAs) [41], [42] and quantum neural networks (QNNs) [43], [44]. Although these methods target different applications, they share a common hybrid mechanism: the quantum processor is responsible for state preparation and measurement, while the classical processor updates and optimizes parameters so that the cost function is minimized. Among them, QNNs use quantum circuits to process feature mappings and combine them with classical optimization for iterative training [43]. Studies show that QNNs possess universal approximation properties and can approximate continuous functions within certain error bounds [44]. Their advantage in effective dimension may also enable faster training than classical networks [45], although barren plateaus remain an important challenge [46]. QNN architectures are increasingly diverse. Quantum convolutional neural networks (QCNNs), for example, have shown potential in image classification and quantum many-body physics [47]. Kyriienko et al. [48] developed QNNs for solving PDEs and used the Navier-Stokes equations to compute density, temperature and velocity distributions in a converging-diverging nozzle. Larki et al. [49] implemented efficient high-accuracy data classification using a quantum Fourier transform based QNN. Tian et al. [50] reviewed recent advances in QNNs for generative learning from a machine-learning perspective and analysed the intrinsic links among different models. Peng et al. [51] proposed HyQ2, a QNN combining graph embedding and variational quantum circuits, and achieved high accuracy in 5G vulnerability detection. Zheng et al. [52] used a quantum graph convolutional network for graph-structured data and verified the model's learning ability, generalization and robustness. Rao et al. [53] applied QNNs to prediction of carbon dioxide sequestration in saline aquifers, and Cho et al. [54] developed a hybrid QNN for quantitative structure-property relationship modelling of CO2-capturing amines.

Despite their advantages over traditional neural networks in high-dimensional feature mapping and complex nonlinear representation, QNNs remain essentially data-driven models if physical constraints are absent. Without such constraints, QNNs can suffer from insufficient interpretability, poor training stability and limited generalization, especially in PDE solution and scientific-computing tasks. A natural idea is therefore to combine physics-informed constraints with QNNs and construct quantum physics-informed neural networks (QPINNs), in which PDEs, boundary conditions and initial conditions are introduced into the loss function to improve quantum-network performance. For example, Trahan et al. [55] applied QPINNs to a spring-mass system and the Poisson equation; Xiao et al. [56] used QPINNs to solve forward and inverse problems for specific PDEs; and Berger et al. [57] applied QPINNs to the Poisson, Burgers and Navier-Stokes equations. However, purely quantum neural networks theoretically require more layers than hybrid quantum-classical networks, which substantially raises hardware-implementation requirements and makes them more susceptible to noise. In addition, hybrid quantum-classical neural networks can exploit the advantages of classical networks and realize nonlinear complexity at much lower cost than fully quantum networks, thereby enhancing nonlinear fitting ability.

Consequently, hybrid quantum-classical physics-informed neural networks (QCPINNs) have been developed. Fernandez et al. [58] integrated quantum layers into PINNs to construct QCPINN and preliminarily tested its performance for the wave equation; Dehaghani et al. [59] applied QCPINN to quantum optimal-control problems; Trahan et al. [55] and Leong et al. [60] used QCPINNs to solve the one-dimensional Burgers equation and two-dimensional inviscid compressible flow, respectively; and Farea et al. [61] applied QCPINNs to the Helmholtz equation, wave equation, Klein-Gordon equation and other PDEs. Numerical experiments by Trahan et al. [55] and Farea et al. [61] showed that, compared with classical PINNs, QCPINNs can use quantum superposition and entanglement to construct high-dimensional feature spaces. This makes it possible to solve PDEs with far fewer parameters than traditional PINNs and substantially improves parameter efficiency. Song et al. [62] implemented a QCPINN solution of the Navier-Stokes equations on noisy quantum hardware, demonstrating the feasibility of the framework for complex flow modelling. Hegde et al. [63] used QCPINN as an electrostatic-field Poisson solver. Rao et al. [64] extended QCPINN to several representative PDEs in porous-media flow and found that it achieved higher computational accuracy than traditional PINNs. Lantigua et al. [65] proved that QCPINNs not only retain universal approximation properties but can also actively alleviate barren plateaus and ensure gradient stability during training. QCPINN therefore provides a breakthrough solution to key limitations of QPINN by combining classical preprocessing, quantum-core computation and classical postprocessing, enabling a deep integration of quantum advantages and physical constraints.

Here, we present QCPIKAN, the first hybrid quantum-classical physics-informed Kolmogorov-Arnold network, by integrating KAN layers into QCPINN. This new architecture addresses the inherent limitations of MLPs in function representation, parameter efficiency and interpretability. Critically, we provide rigorous theoretical derivation and approximation analysis, demonstrating that QCPIKAN lifts the error convergence of high-frequency components from algebraic to exponential order. This fundamental improvement substantially suppresses numerical dispersion, a long-standing challenge in simulating fields with steep gradients. We systematically benchmark the performance of QCPIKAN against QCPINN on three canonical porous media flow problems: pressure diffusion, component advection-diffusion, and coupled pressure-saturation flow for two-phase systems. Our comparisons focus particularly on computational accuracy within local high-gradient regions, which are representative of realistic seepage scenarios.

The rest of the paper is structured as follows. Section II details the complete QCPIKAN architecture alongside the theoretical analysis based on approximation theory. This section covers the Chebyshev-polynomial KAN preprocessor, DV-Circuit quantum core, KAN postprocessor and composite loss function. Section III conducts comprehensive numerical tests on classic porous media flow equations, with quantitative assessments covering loss convergence, local peak error, front localization accuracy and temporal evolution of physical fields. Finally, Section IV summarizes the key findings of this work.

## Methods

### QCPIKAN: quantum-classical physics-informed Kolmogorov-Arnold network

The quantum-classical physics-informed KAN proposed in this work, termed QCPIKAN, is a new solution framework constructed by systematically replacing the classical MLP layers in QCPINN architecture with KAN layers. Therefore, as shown in Fig. 1, QCPIKAN consists of three serial components: (1) a KAN preprocessor, which uses Chebyshev polynomials as basis functions to map space-time coordinates to quantum-circuit inputs $\mathbf{X}_q$ ; (2) a DV quantum core, which uses angle encoding and parameterized quantum circuits to realize Hilbert-space feature mapping; and (3) a KAN postprocessor, which decodes the quantum measurement results $\mathbf{Y}_q$ into physical-field variables. It should be noted that the KAN implementation adopted in this work does not directly use the spline-parameterized form of the original KAN proposed by Liu et al. [16], but instead uses a lightweight implementation based on Chebyshev polynomials. This design is simple, numerically stable and well compatible with the quantum core module.

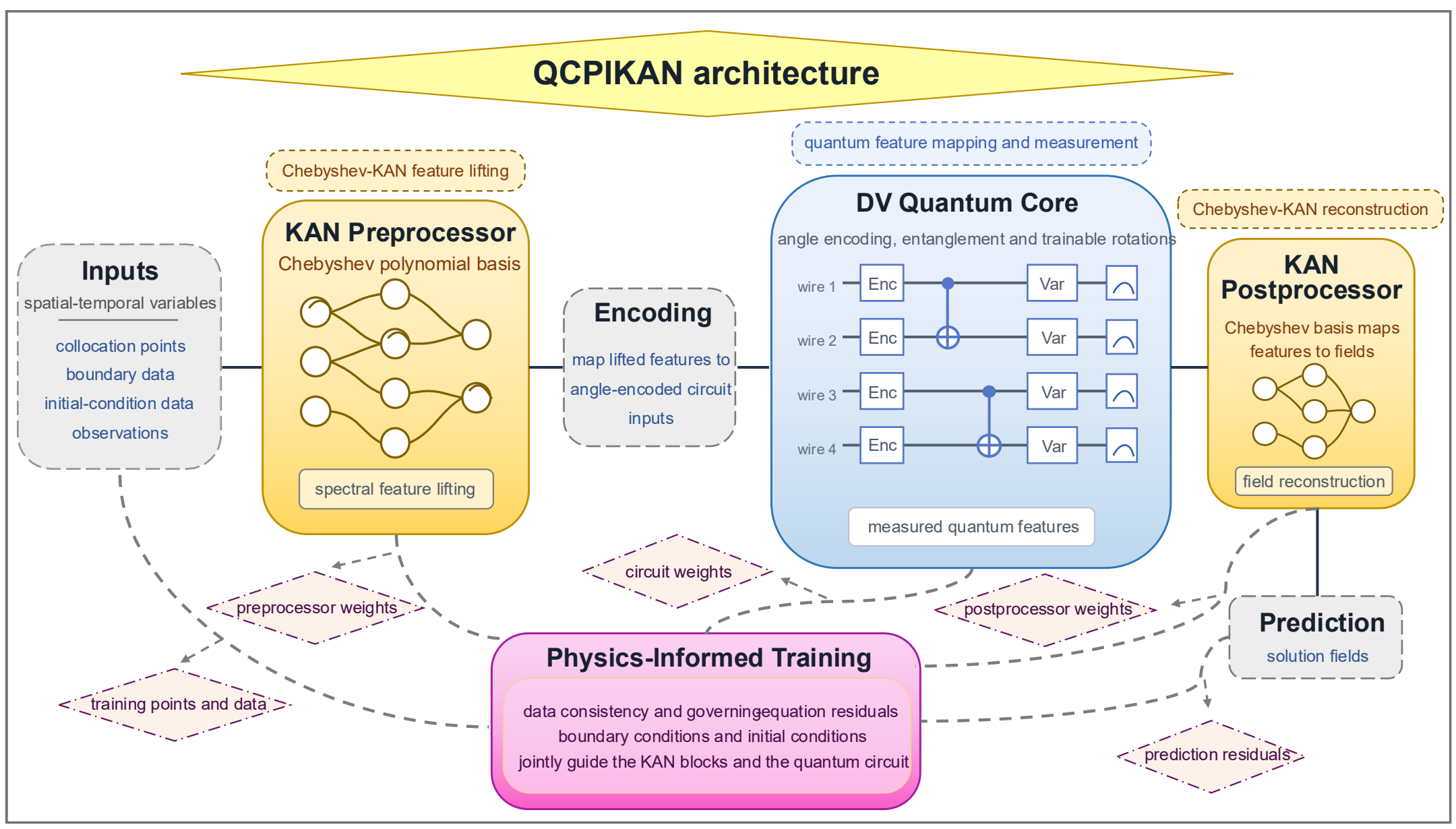

Fig. 1. Proposed QCPIKAN architecture.

**KAN preprocessor**. The core idea of KAN originates from the Kolmogorov-Arnold Network architecture proposed by Liu et al. [16]. Its main idea is to replace fixed activation functions on the nodes of traditional neural networks with learnable univariate functions on the edges, thereby enhancing the ability of the network to represent complex nonlinear mappings. Unlike traditional MLPs, which mainly rely on nodal nonlinearities, KAN realizes a more flexible function-approximation mechanism by adaptively learning edge functions. In this work, Chebyshev polynomials of the first kind are selected as the basis functions, defined on the $[-1,1]$ interval as

$$T_n(x) = \cos(n\arccos x), \tag{1}$$

and satisfying the recurrence relation

$$T_{n+1}(x) = 2xT_n(x) - T_{n-1}(x) . \tag{2}$$

This set of polynomials is orthogonal on the interval $[-1,1]$ with respect to the corresponding weight function,

$$w(x) = \frac{1}{\sqrt{1-x^2}} , \tag{3}$$

and therefore has excellent spectral-approximation properties. For functions analytic inside a Bernstein ellipse, the Chebyshev approximation error can converge exponentially as the polynomial order increases [66]. Based on this property, Sidharth et al. [67] further proposed the ChebyshevKAN structure, in which Chebyshev polynomials replace the B-spline basis functions in the traditional KAN, thereby improving training efficiency and numerical stability.

Following these studies, the univariate function connecting the $i$-th neuron in layer $l$ to the $j$-th neuron in layer $l+1$ of a KAN layer is $T_{i,j}^{l}$ parameterized as a linear combination of Chebyshev polynomials of order $K$:

$$T_{i,j}^{l}(x) = \sum_{k=0}^{K} c_{i,j,k}^{l} \cdot T_k\left(\tanh(x)\right) , \tag{4}$$

where the coefficients $c_{i,j,k}^{l} \in \mathbb{R}$ are learnable. Any real-valued input is nonlinearly compressed by $\tanh(\cdot)$ to match the natural definition domain $(-1,1)$ of the Chebyshev polynomials. The coefficients are initialized from a normal distribution with mean zero and a prescribed standard deviation $1/\left(d_{\text{in}} \times (K+1)\right)$ so that the activation values of each neuron have a reasonable scale at the early stage of forward propagation. The computational process used in this work is logically consistent with the forward propagation of the ChebyKANLayer in the official ChebyKAN implementation: the input is first compressed by the hyperbolic tangent to the target interval $[-1,1]$, then Chebyshev basis functions are computed in angular form, and finally the trainable coefficients are used for weighted combination to obtain the output.

The preprocessor adopts a KAN structure with one hidden layer and contains two KAN transformations: $\mathbb{R}^{d_{\text{in}}} \to \mathbb{R}^{d_h} \to \mathbb{R}^{d_q}$. Let the input be the coordinate vector $\mathbf{X} \in \mathbb{R}^{d_{\text{in}}}$, the number of hidden neurons be $d_h$, the number of qubits be $d_q$ and the polynomial order be $K$ .

The first KAN transformation maps the input to hidden-layer features $\mathbf{h}_1 \in \mathbb{R}^{d_h}$ :

$$h_{1,i} = \sum_{j=1}^{d_{\text{in}}} T_{i,j}^{(1)}\left(X_j\right) = \sum_{j=1}^{d_{\text{in}}} \sum_{k=0}^{K} c_{i,j,k}^{(1)} \cdot \cos\left(k \cdot \arccos\left(\tanh\left(X_j\right)\right)\right), i = 1, 2, \ldots, d_h , \tag{5}$$

where the $\tanh\left(X_j\right)$ inner nonlinear compression maps the $j$-th input component to the Chebyshev domain $(-1,1)$; the Chebyshev definition , given by $\cos\left(k \cdot \arccos(\cdot)\right)$, is then used to compute the $k$-th basis-function value corresponding to this input; the learnable coefficients $c_{i,j,k}^{(1)}$ weight the $K+1$ basis functions; and the contributions from all $d_{\text{in}}$ input components are summed to obtain the value of the $i$-th hidden neuron.

The second KAN transformation maps the hidden-layer features to the input of the quantum circuit $\mathbf{X}_q \in \mathbb{R}^{d_q}$:

$$x_{q,l} = \sum_{i=1}^{d_h} T_{l,i}^{(2)}\left(h_{1,i}\right) = \sum_{i=1}^{d_h}\sum_{k=0}^{K} c_{l,i,k}^{(2)} \cdot \cos\left(k \cdot \arccos\left(\tanh\left(h_{1,i}\right)\right)\right), \, l = 1, 2, \ldots, d_q \, . \tag{6}$$

All learnable coefficients in the two transformations, $\left\{c_{i,j,k}^{(1)}, c_{l,i,k}^{(2)}\right\}$, are collected together and denoted as $\theta_1$, and the overall preprocessor map is denoted as $\mathbf{X}_q = f_{\text{pre}}\left(\mathbf{X}; \boldsymbol{\theta}_1\right)$.

Compared with an MLP preprocessor, each connection in the KAN preprocessor uses an adaptive combination of $K+1$ Chebyshev basis functions as the activation function, rather than a single fixed pointwise nonlinear function. Under a comparable total number of parameters, this design enables the preprocessor to learn richer input-output mappings and provides a more informative initial feature representation for the subsequent quantum circuit.

**Quantum core.** The quantum core maps the classical features output by the KAN preprocessor to a high-dimensional Hilbert space and uses quantum resources to enhance feature representation [68], [69]. This work follows the DV-Circuit quantum-core framework [61], which supports multiple quantum-circuit topologies for different quantum feature-mapping tasks. The quantum-core design refers to Rao et al. [64], who used typical Cascade, Cross-mesh and Alternate topologies to apply QCPINN to representative seepage equations. Here, the Cross-mesh topology with a fully connected entanglement structure is taken as an example to describe the construction of the quantum core and the process of mapping classical features to quantum states. The classical input first passes through the KAN preprocessor. Then, by angle encoding, each component of the preprocessor output $\mathbf{X}_q \in \mathbb{R}^{d_q}$ is used as a rotation angle, and RX rotation gates are applied to the initial computational basis states $|0\rangle^{\otimes d_q}$ of the $d_q$ qubits to encode classical information into the quantum state:

$$|\psi_{\text{enc}}\rangle = \prod_{j=1}^{d_q} R_X(x_{q,j})\,|0\rangle^{\otimes d_q} \, , \tag{7}$$

$$R_X(\theta) = e^{-i\theta X/2} = \begin{pmatrix} \cos\frac{\theta}{2} & -i\sin\frac{\theta}{2} \\ -i\sin\frac{\theta}{2} & \cos\frac{\theta}{2} \end{pmatrix}. \tag{8}$$

The parameterized quantum circuit of the Cross-mesh topology is composed of $L$ circuit layers ($L=1$ in this work). Each layer contains three sublayers: a pre-rotation layer, which applies $R_Z$ gates with trainable parameters $\eta_{l,j}$ and $R_X$ gates with trainable parameters $\gamma_{l,j}$ to each qubit; a fully connected entanglement layer, which applies controlled-$R_Z$ gates with trainable parameters $\delta_{l,km}$ to all pairs of qubits $(k,m)$ where $k \neq m$; and a post-rotation layer, which applies $R_X$ gates with trainable parameters $\alpha_{l,j}$ and $R_Z$ gates with trainable parameters $\beta_{l,j}$ to each qubit. All rotation angles are trainable parameters, and the quantum state evolves through $L$ layers to obtain the final quantum state:

$$|\psi_{\text{final}}\rangle = \prod_{l=1}^{L}\left[\prod_{j=1}^{d_q} R_X(\gamma_{l,j}) R_Z(\eta_{l,j}) \prod_{1\le k \neq m \le d_q} \text{CRZ}(\delta_{l,km}) \prod_{j=1}^{d_q} R_Z(\beta_{l,j}) R_X(\alpha_{l,j})\right] |\psi_{\text{enc}}\rangle \, . \tag{9}$$

Here, the trainable parameters are collected in the quantum parameter set $\boldsymbol{\theta}_2=\{\alpha_{l,j},\beta_{l,j},\gamma_{l,j},\eta_{l,j},\delta_{l,km}\}$. The Pauli-Z expectation value of each qubit is measured to obtain the output vector of the quantum core $\mathbf{Y}_q\in\mathbb{R}^{d_q}$:

$$y_{q,j}=\langle\psi_{\text{final}}|Z_j|\psi_{\text{final}}\rangle,\ Z=|0\rangle\langle 0|-|1\rangle\langle 1|,\ j=1,\ldots,d_q. \tag{10}$$

The detailed structures of the Cascade and Alternate topologies are described in the work of Farea et al. [61]. The topology used in each numerical example in this work is specified together with the corresponding problem setup in the results section.

**KAN postprocessor**. The postprocessor decodes the quantum measurement results $\mathbf{Y}_q\in\mathbb{R}^{d_q}$ into the target physical-field variables. Symmetric to the preprocessor, the postprocessor also adopts a KAN structure with one hidden layer. The first KAN layer maps the quantum measurement vector $\mathbf{Y}_q$ to hidden features $\mathbf{h}_2\in\mathbb{R}^{d_h}$:

$$h_{2,i}=\sum_{j=1}^{d_q}T_{i,j}^{(3)}\left(y_{q,j}\right)=\sum_{j=1}^{d_q}\sum_{k=0}^{K}c_{i,j,k}^{(3)}\cdot\cos\left(k\cdot\arccos\left(\tanh\left(y_{q,j}\right)\right)\right),\ i=1,\ldots,d_h. \tag{11}$$

The second KAN layer maps the hidden features to the final output $Y\in\mathbb{R}$, such as pressure, saturation or component concentration:

$$Y=\sum_{i=1}^{d_h}T_{1,i}^{(4)}\left(h_{2,i}\right)=\sum_{i=1}^{d_h}\sum_{k=0}^{K}c_{1,i,k}^{(4)}\cdot\cos\left(k\cdot\arccos\left(\tanh\left(h_{2,i}\right)\right)\right). \tag{12}$$

All learnable coefficients $\{c_{i,j,k}^{(3)},c_{1,i,k}^{(4)}\}$ of the postprocessor are collected $\boldsymbol{\theta}_3$ together, and the postprocessor is denoted accordingly as $Y=f_{\text{post}}\left(\mathbf{Y}_q;\boldsymbol{\theta}_3\right)$.

**Loss function.** By cascading the preprocessor, quantum core and postprocessor, the overall QCPIKAN mapping can be written as

$$\mathbf{Y}=f_{\text{QCPIKAN}}\left(\mathbf{X};\boldsymbol{\Theta}\right),\ \boldsymbol{\Theta}=\{\boldsymbol{\theta}_1,\boldsymbol{\theta}_2,\boldsymbol{\theta}_3\}. \tag{13}$$

To ensure that the network output satisfies the governing physical laws, the following composite loss function is constructed:

$$L(\boldsymbol{\Theta})=\lambda_1 L_{\text{PDE}}(\boldsymbol{\Theta})+\lambda_2 L_{\text{BC}}(\boldsymbol{\Theta})+\lambda_3 L_{\text{IC}}(\boldsymbol{\Theta}), \tag{14}$$

where $L_{\text{PDE}}$ is the PDE residual loss on interior collocation points, which consists of the mean-square residual of the governing PDE:

$$L_{\text{PDE}}=\frac{1}{N_{\text{in}}}\sum_{i=1}^{N_{\text{in}}}\left[\mathcal{R}_{\text{PDE}}^2\left(\mathbf{x}_i,t_i\right)\right]. \tag{15}$$

$L_{\text{BC}}$ is the boundary-condition loss, which includes the deviation between the network prediction and the prescribed boundary conditions on Dirichlet and Neumann boundaries:

$$L_{\text{BC}}=\frac{1}{N_{\text{D}}}\sum_{i=1}^{N_{\text{D}}}\left[p\left(\mathbf{x}_i,t_i\right)-p_{\text{D}}\left(\mathbf{x}_i,t_i\right)\right]^2+\frac{1}{N_{\text{N}}}\sum_{j=1}^{N_{\text{N}}}\left[\nabla p\left(\mathbf{x}_j,t_j\right)\cdot\mathbf{n}-q_{\text{N}}\left(\mathbf{x}_j,t_j\right)\right]^2, \tag{16}$$

where $N_{\mathrm{D}}$ and $N_{\mathrm{N}}$ are the numbers of collocation points on the Dirichlet and Neumann boundaries, $p_{\mathrm{D}}$ is the prescribed boundary pressure, and $q_{\mathrm{N}}$ is the prescribed boundary flux are defined for the corresponding boundary segments. For the saturation field, the boundary-condition loss is constructed in the same form.

$L_{\mathrm{IC}}$ is the initial-condition loss, which is activated only for transient problems:

$$L_{\mathrm{IC}} = \frac{1}{N_{\mathrm{IC}}} \sum_{i=1}^{N_{\mathrm{IC}}} \left[ \mathbf{Y}\left(\mathbf{x}_i, 0\right) - \mathbf{Y}_0\left(\mathbf{x}_i\right) \right]^2 . \tag{17}$$

All loss terms are measured by mean-square error. Interior, boundary and initial collocation points are generated by Latin hypercube sampling. All spatial and temporal derivatives in the model are computed by PyTorch automatic differentiation. The weights of the loss terms, $\lambda_1$, $\lambda_2$, $\lambda_3$ are all set to one in the experiments of this work. Training is uniformly performed using the Adam optimizer, and the specific learning rate, learning-rate scheduling strategy and regularization settings for each example are given in the corresponding model-configuration table.

For coupled problems that require simultaneous prediction of multiple physical variables, this work constructs multiple independent QCPIKAN subnetworks with the same architecture to approximate each physical variable separately. The loss terms of each subnetwork are computed independently and then summed as the total loss. For single-physical-variable problems, a single-network architecture is used.

## Results

### Approximation-error analysis of QCPIKAN

To further reveal the advantage of QCPIKAN in dealing with discontinuous or high-gradient physical fields, this section gives a supplementary analysis from three aspects: the approximation order of high-frequency components, the error control of the gradient field and the front-localization accuracy. Let the exact field $u$ be decomposed into low-frequency part $u_{low}$ and high-frequency part $u_{high}$, where the support of $u_{high}$, denoted as $\Gamma = \operatorname{supp}\left(u_{\mathrm{high}}\right) \subset \Omega$, consists of a finite number of smooth curves (fronts). Furthermore, $u_{high}$ has a first-kind jump discontinuity on $\Gamma$, that is, for any $\boldsymbol{x} \in \Gamma$, it holds that: $\lim_{\varepsilon \to 0^+} \left| u_{\mathrm{high}}\left(\boldsymbol{x} + \varepsilon \boldsymbol{n}\right) - u_{\mathrm{high}}\left(\boldsymbol{x} - \varepsilon \boldsymbol{n}\right) \right| \geq \delta > 0$, where $\boldsymbol{n}$ is the unit normal vector. For a model $\mathcal{M}$, its high-frequency approximation error is defined as

$$E_{\mathrm{high}}\left(\mathcal{M}\right) = \inf_{h \in \mathcal{F}_{\mathcal{M}}} u_{\mathrm{high}} - h_{\mathrm{high}\, L^2(\Omega)} , \tag{18}$$

where $\mathcal{F}_{\mathcal{M}}$ is mapping space of the model $\mathcal{M}$, and $h_{\mathrm{high}}$ denotes the projection of the model output field $h$ onto the support subspace of $u_{high}$ in the $L^2$ sense. If there exists a constant $C > 0$ and a convergence-rate function $\rho(N)$ (where $N$ is the total number of optimizable parameters) such that $E_{\mathrm{high}}\left(\mathcal{M}\right) \leq C\rho(N)$, then $\rho(N)$ is referred to as the high-frequency convergence order of the model.

Under the above setting, the following convergence-order estimate for approximating high-frequency components can be obtained. Under the constraint that the quantum-circuit structure, quantum parameters and total number of model parameters $N$ are identical, and denoting $N$ as the number of optimizable parameters in the classical subnetworks (or the corresponding width/depth scale), there exist positive constants $c$, $C_1$, $C_2$, $\alpha > 0$ (with $\alpha < 1$) independent of this scale $N$ such that

$$E_{\text{high}}\left(\text{QCPIKAN}\right) \le C_1 e^{-cN}, \tag{19}$$

$$E_{\text{high}}\left(\text{QCPINN}\right) \ge C_2 N^{-\alpha} \quad \left( or = O\left(N^{-\alpha}\right)\right). \tag{20}$$

The proof of this conclusion is as follows. First, we consider the error behavior of QCPIKAN. The front end of QCPIKAN employs Chebyshev-KAN, whose core is $M$-th order Chebyshev polynomials spectral approximation. Consider a function $u_{\text{high}}$ defined on $\left[-1,1\right]^2$, which can be generalized to a general bounded domain $\Omega$ via an affine transformation. Its Chebyshev expansion is

$$u_{\text{high}}\left(\boldsymbol{x}\right) = \sum_{k_1,k_2=0}^{\infty} \hat{u}_{k_1 k_2} T_{k_1}\left(x_1\right) T_{k_2}\left(x_2\right), \tag{21}$$

where the basis functions are Chebyshev polynomials of the first kind, and the expansion coefficients are the corresponding Chebyshev coefficients $\hat{u}_{k_1 k_2}$. According to spectral method theory, if the function $u_{\text{high}}$ is analytic on the domain $\Omega \backslash \Gamma$ except for a finite number of jump discontinuities on $\Gamma$, its Chebyshev coefficients satisfy an exponential decay estimate [70]: there exists a constant $0 < r < 1$ such that $|\hat{u}_{k_1 k_2}| \le C r^{\max(k_1,k_2)}$. Taking the truncation order $M = \lfloor \gamma N \rfloor$ (where $\gamma > 0$, since the number of parameters in Chebyshev-KAN is proportional to $M$), the high-frequency approximation error after applying the truncation operator $T_k$ satisfies the stated exponential bound.

$$\left\| u_{\text{high}} - f_{\text{high}}^{(\text{pre})} \right\|_{L^2} \le C e^{-cM} = O\left(e^{-cN}\right). \tag{22}$$

The subsequent quantum core applies a unitary transformation $\mathcal{U}\left(\boldsymbol{\theta}\right)$ on the features in a $2^{N_q}$-dimensional Hilbert space, and this transformation preserves the $L^2$ norm. Quantum entanglement can further sharpen the gradient features near the front [71], but it does not destroy the exponential accuracy already obtained by the Chebyshev-KAN component. Therefore, the final approximation error of QCPIKAN retains the exponential behavior $E_{\text{high}}\left(\text{QCPIKAN}\right) = O\left(e^{-cN}\right)$.

On the other hand, the front end of QCPINN is an MLP. Let $g_{\text{high}}$ denote the high-frequency component output by the MLP. Classical results [72] show that, for functions with finite Barron norm, the approximation error of a shallow network is $O\left(N^{-1/2}\right)$. However, since $u_{\text{high}}$ is discontinuous on $\Gamma$, its Barron norm is unbounded and it belongs to a rougher function class. According to Neural Tangent Kernel (NTK) theory [73], MLPs exhibit spectral bias during training: low-frequency components converge at an exponential rate, whereas high-frequency components converge only algebraically. Specifically, for discontinuous functions, the decay of high-frequency Fourier coefficients in the MLP representation is limited to $O\left(|\omega|^{-1}\right)$, which leads to an algebraic high-frequency error even after optimal training. More specifically, the error satisfies: $\left\| u_{\text{high}} - g_{\text{high}} \right\|_{L^2} = O\left(N^{-\alpha}\right)$, $\alpha < 1$. More refined analyses indicate that the exponent $\alpha$ lies within the range $\left[1/2, 1/3\right]$ [74]. Since the high-frequency information input to the quantum core has already been severely attenuated by the MLP stage, equivalent to low-pass filtering, the quantum core cannot recover the lost high-frequency details. Hence the overall convergence of $E_{\text{high}}(\text{QCPINN})$ remains limited to algebraic order. Combining the results from these two aspects yields Eq. (19) and Eq. (20).

The error control can be further expressed from the perspectives of gradient fields and front position. Let the true gradient field $u$ be denoted by $\nabla u$ and the gradient field of the model predicted field $v$ be $\nabla v$ (where the gradient at discontinuities is understood in the distributional sense). Define $\Gamma = \{\boldsymbol{x} \in \Omega : |\nabla u(\boldsymbol{x})| \text{ is unbounded or } |\nabla u| \text{ exceeds } T\}$ as the front set. Then there exist positive constants $C_1$, $C_2$, $c$, $\alpha$, independent of the model, such that:

$$\|\nabla u - \nabla v_{\text{QCPIKAN}}\|_{L^2(\Omega)} \le C_1 e^{-cN}, \tag{23}$$

$$\|\nabla u - \nabla v_{\text{QCPINN}}\|_{L^2(\Omega)} \ge C_2 N^{-\alpha} \ (\text{or} = O(N^{-\alpha})). \tag{24}$$

In addition, defining the model-predicted front as $\hat{\Gamma}_{\mathcal{M}} = \{\boldsymbol{x} \in \Omega : |\nabla v_{\mathcal{M}}(\boldsymbol{x})| > T\}$ (using the same threshold $T$), the Hausdorff distance satisfies

$$d_H\left(\Gamma, \hat{\Gamma}_{\text{QCPIKAN}}\right) = O\left(e^{-cN}\right), d_H\left(\Gamma, \hat{\Gamma}_{\text{QCPINN}}\right) = O\left(N^{-\alpha}\right). \tag{25}$$

These estimates can be formally derived based on $L^2$ approximation results. Consider the distributional gradient of $u$, denoted as $\nabla u = \nabla u_{\text{low}} + \nabla u_{\text{high}}$, where $\nabla u_{\text{high}}$ contains a Dirac-measure component on $\Gamma$.Applying elliptic regularity estimates or inverse inequalities show that, on compact subsets away from $\Gamma$, the $L^2$ error can already control the $H^1$ error. Near $\Gamma$ the front, gradient-error estimates for spectral approximation of discontinuous functions [70] give the corresponding bound.

$$\|\nabla u - \nabla v_{\text{QCPIKAN}}\|_{L^2(\Omega)} \le C\|u - v_{\text{QCPIKAN}}\|_{L^2(\Omega)} + O\left(e^{-cN}\right) = O\left(e^{-cN}\right). \tag{26}$$

For MLPs, the corresponding estimate can only give algebraic order. For the front position, threshold-detection stability implies that, when $\|\nabla u - \nabla v\|_{L^2}$ is sufficiently small, the Hausdorff distance between $\Gamma$ the region where the modulus of $\nabla v$ exceeds $T$ is controlled by this error, thereby obtaining Eq. (25). More detailed arguments can be found in the inverse inequalities for spectral methods in [70] and the quantitative analysis of MLP frequency bias in [74].

In summary, when the physical variable contains discontinuities or high-gradient distributions, the overall approximation error, gradient error and front-localization accuracy of QCPIKAN are exponentially better than the algebraic-order behavior of QCPINN. Therefore, in seepage problems such as oil-water two-phase displacement, QCPIKAN can significantly reduce numerical dispersion and capture the front position more accurately.

**Numerical experiments**

The numerical experiments in this work are carried out on the basis of the three representative seepage equations tested by Rao et al. [64], and the best-performing topology in each example is selected to construct the quantum core. The physical characteristics of the three examples increase progressively in complexity, covering the main problem types in reservoir numerical simulation from single-phase to two-phase flow, from steady-state to transient behavior, and from a single mechanism to coupled multiple mechanisms. In each example, QCPIKAN and QCPINN are compared under the same experimental conditions.

All three examples use the following unified configuration: the optimizer is Adam, combined with the ReduceLROnPlateau learning-rate scheduler with a decay factor of 0.9 and a patience of 1000; the total number of training epochs is 20000; the number and distribution of training sampling points are kept the same for the two models; the loss weights are set to one ( $\lambda_1 = \lambda_2 = \lambda_3 = 1$ ); angle encoding is used; and the Chebyshev polynomial order in QCPIKAN is set to 3. The differentiated configurations for each example, including quantum-circuit topology, number of qubits, number of quantum layers, classical-network structure, total number of trainable parameters, learning rate and batch size, are given in the corresponding model-configuration tables.

*Example 1: heterogeneous single-phase flow*

In this example, the reservoir pressure $p(x,y)$ distribution is governed by the Darcy equation for single-phase flow. Its mathematical form in a heterogeneous reservoir is

$$\nabla \cdot \left( \frac{k(x,y)}{\mu} \nabla p(x,y) \right) = 0, \quad (27)$$

where $k(x,y) = (100 + y)mD$ is the spatially dependent absolute permeability, and $\mu = 1\text{cp}$ is a rectangular reservoir prescribed. The computational domain is a rectangular reservoir $[0,100] \times [0,100]\text{m}^2$. The boundary conditions are a fixed pressure of 5 MPa on the upper boundary and 10 MPa on the lower boundary, while the left and right boundaries are impermeable closed boundaries. QCPIKAN and QCPINN are employed to approximate Eq. (27). The network input is the spatial coordinate $(x,y)$, and the output is the pressure $p(x,y)$. By imposing the PDE and boundary conditions, the network learns the pressure distribution over the whole spatial domain. It should be noted that the spatial coordinates are normalized to the range [0, 1] for the network models in actual calculations of all examples.

Table 1 presents the distinct hyperparameters of QCPINN and QCPIKAN for the test example, while Fig. 2 illustrates their training loss curves. Since the pressure field bears no gradient along the *x* direction, the governing equation simplifies to an ordinary differential equation solely dependent on the other axis. Its analytical solution derived from boundary conditions serves as the benchmark for error assessment. Figs. 3 and 4 compare the predictive performance of the two models. QCPIKAN yields far smaller local errors near boundaries and high-gradient zones, offering tighter control over peak local errors. At the top 20, 50 and 100 error points, QCPIKAN consistently exhibits smaller maximum errors than QCPINN, verifying its strength in mitigating extreme local deviations. Collectively, QCPIKAN cuts local peak errors and stabilizes predictions in critical areas. It retains decent global accuracy while greatly boosting local prediction reliability, making it more robust for complicated reservoir pressure forecasting.

Table 1. Model-configuration parameters of QCPINN and QCPIKAN in example 1.1

| Configuration | QCPINN | QCPIKAN |
|---|---|---|
| Quantum-circuit topology | Cascade | Cascade |
| Number of qubits | 2 | 2 |

| Quantum layers | 1 | 1 |
|---|---|---|
| Classical-network architecture | [2, 7, 7, 1] | [2, 20, 20, 1] |
| Trainable parameters | 189 | 202 |
| Learning rate | 0.0001 | 0.0001 |
| Batch size | 64 | 64 |

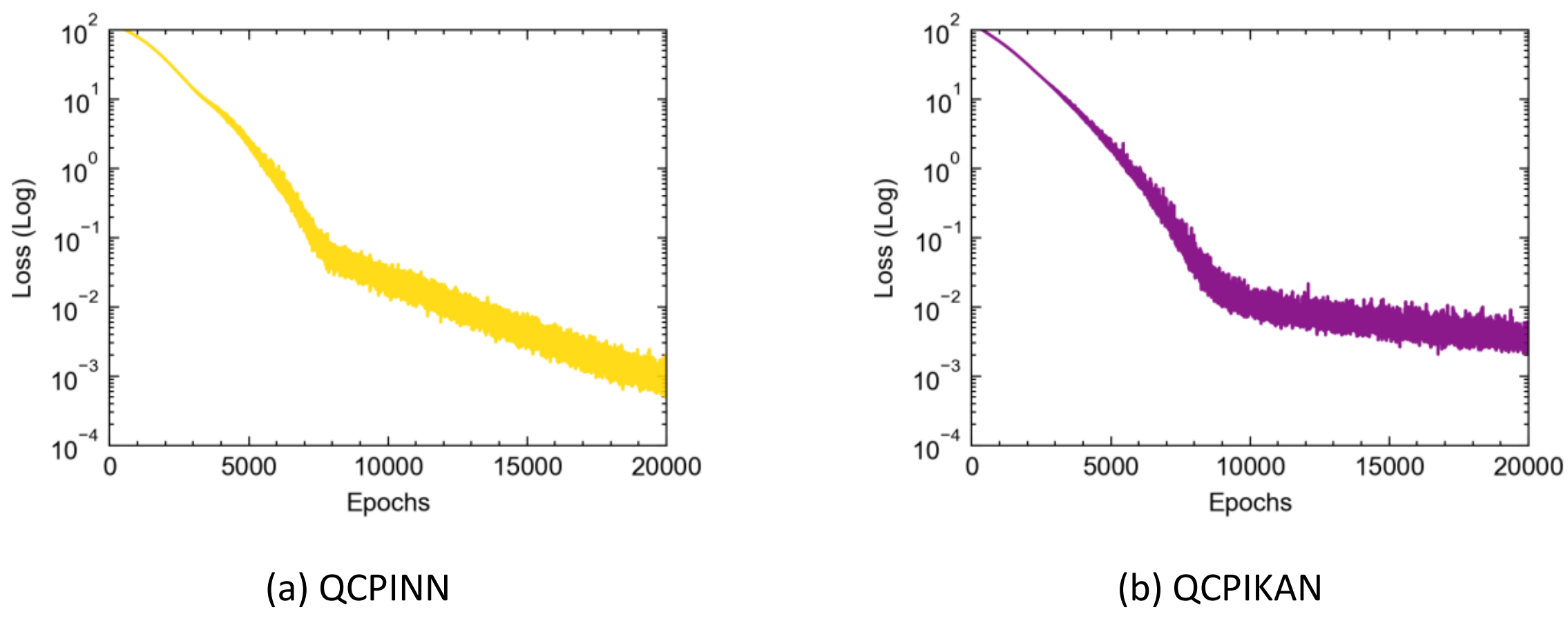

(a) QCPINN (b) QCPIKAN

Fig. 2. Training-loss curves of different methods in example 1.

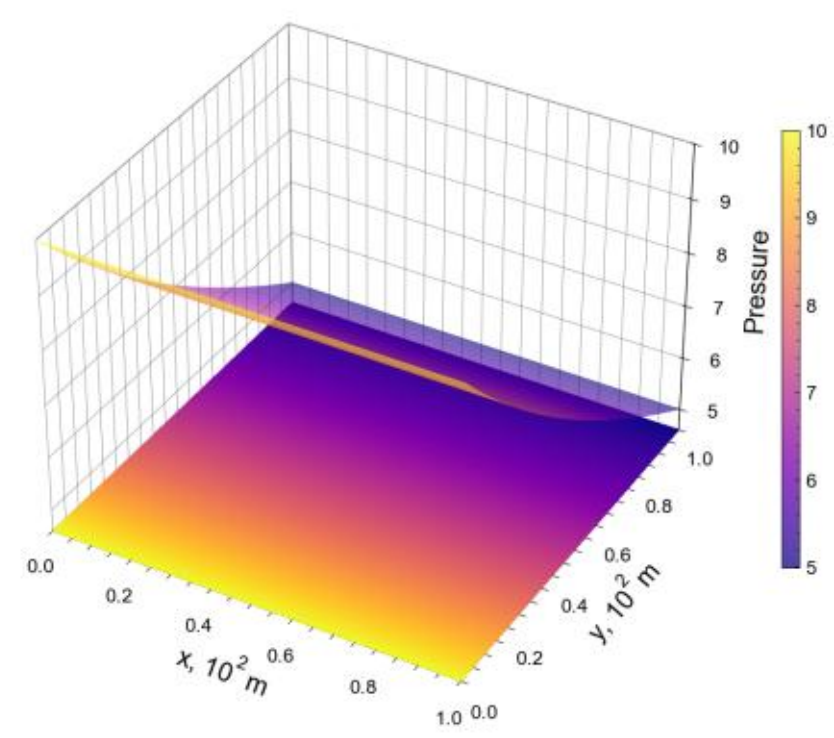

(a) Reference solution

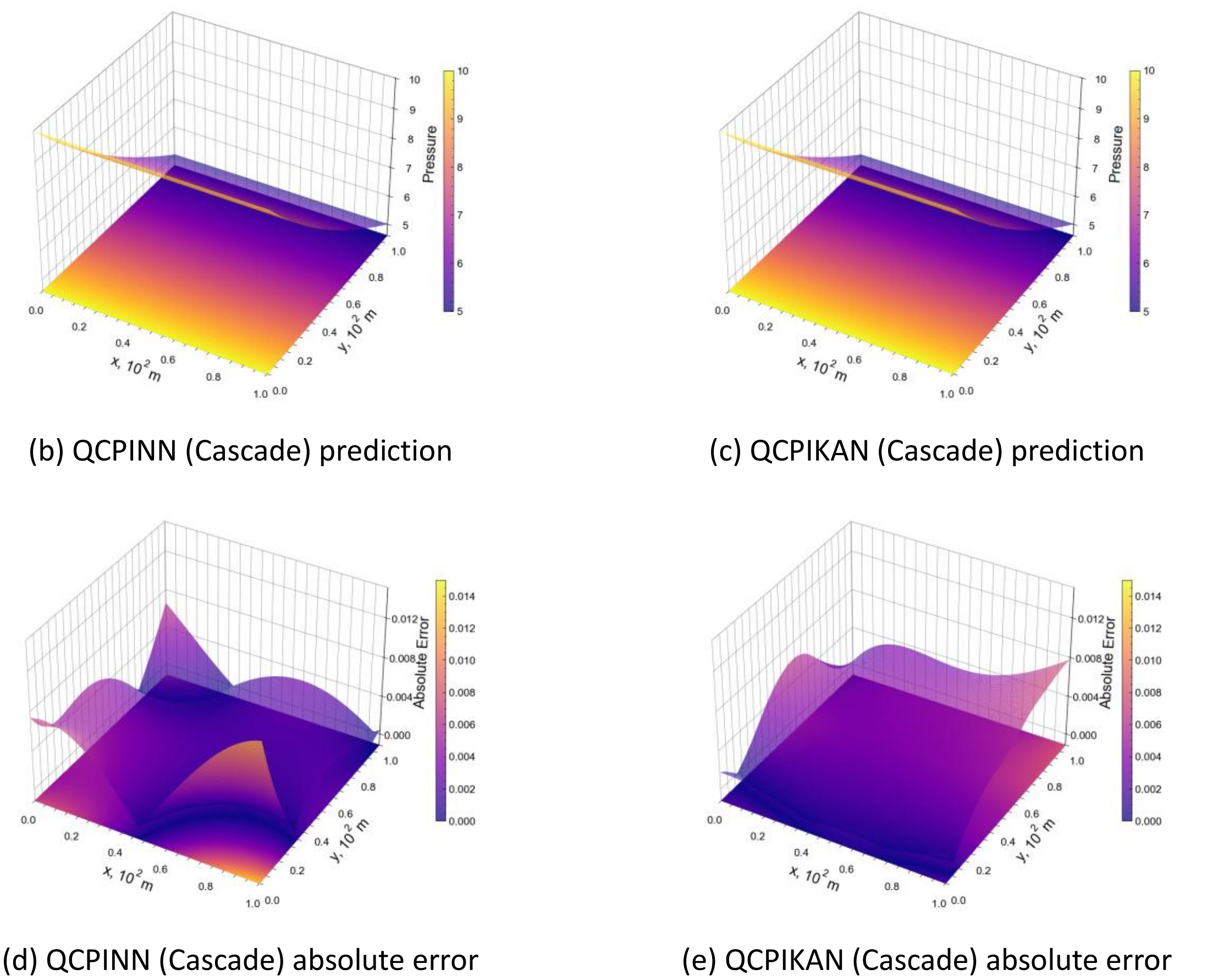

(b) QCPINN (Cascade) prediction (c) QCPIKAN (Cascade) prediction

(d) QCPINN (Cascade) absolute error (e) QCPIKAN (Cascade) absolute error

Fig. 3. Pressure distributions predicted by different methods and the corresponding error distributions in example 1.

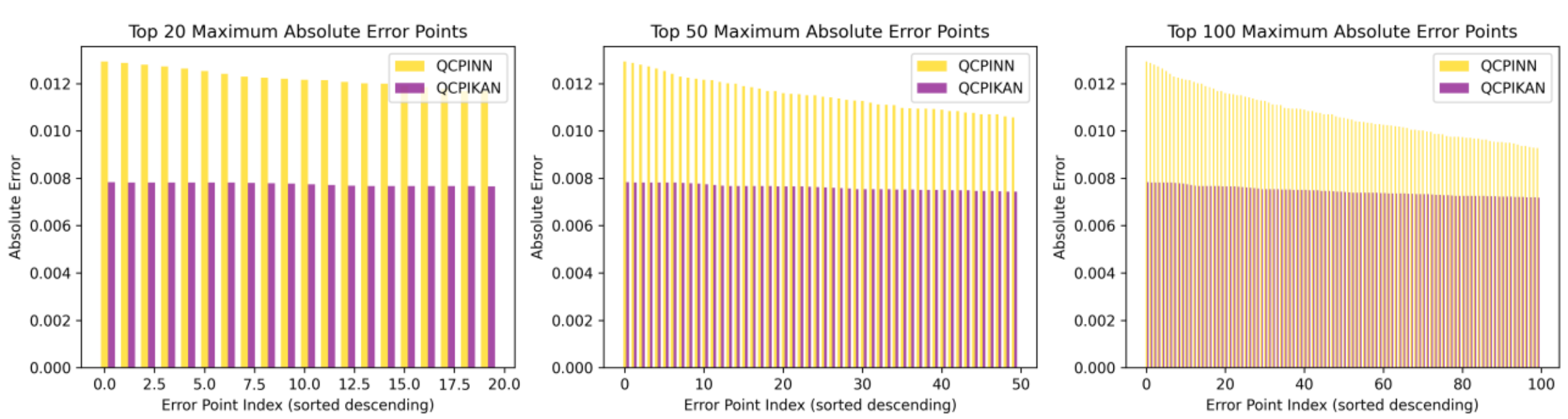

Fig. 4. Bar chart of the Top-K maximum absolute errors.

*Example 2: transient advection, diffusion and adsorption equation*

This example solves the concentration-transport problem in a transient chemical-agent migration process. The governing equation is Eq. (28), which integrates convection, diffusion and adsorption effects:

$$\phi \frac{\partial c}{\partial t} + \rho_b \frac{\partial q}{\partial t} = \nabla \cdot \left( D \nabla c \right) - \nabla \cdot \left( \mathbf{v} c \right), \tag{28}$$

where adsorption follows Henry's law $q = K_d c$ . The porosity is taken as $\phi = 0.3$ , the rock density $\rho_b = 2000\,\mathrm{kg/m^3}$ , the distribution coefficient $K_d = 1.0\times10^{-4}\,\mathrm{m^3/kg}$ , the diffusion coefficient $D = 0.005\,\mathrm{m^2/s}$ , and the Darcy seepage velocity $\mathbf{v} = (0.5, 0)\,\mathrm{m/s}$ . The initial condition is $c(x, y, 0) = 0$ . The left boundary is a constant concentration injection with $c = 1$ , and the right boundary is a constant concentration with $c = 0$ . The upper and lower boundaries are both zero-flux closed boundaries ( $\partial c/\partial y = 0$ ). The network input is the space-time coordinate ( $t, x, y$ ), and the output is the component concentration $c$ .

Table 2 lists the distinct hyperparameters of QCPINN and QCPIKAN for this case. Fig. 5 shows their training loss curves. We adopt the upwind finite difference method to generate reference solutions, using a 200×200 uniform spatial grid and 1000 time steps to resolve transport fronts and concentration gradients for credible benchmark evaluation. Figs. 6 and 7 display absolute error contours of QCPINN and QCPIKAN at two time instants. Errors of both models cluster around concentration fronts due to steep local gradients hindering network fitting. At *t*=0.4 during front propagation, QCPINN presents prominent concentrated high-error zones, while QCPIKAN delivers smoother error fields with smaller high-error areas. At *t*=0.8 with further front movement, QCPINN's intense narrow high-error band becomes more evident, whereas QCPIKAN maintains uniform error distribution and restrains local error spikes near fronts. This proves KAN layers enable spatial local smoothing and mitigate error accumulation in high-gradient zones.

To evaluate the models' capability to capture concentration transport rules, we extract reference, QCPINN and QCPIKAN concentration time series at spatial coordinate (50, 50), with results and corresponding error plotted in Fig. 8. Fig. 8a reveals both models match the general concentration evolution trend, yet QCPIKAN better fits the rapid rising stage with minor phase and amplitude errors. QCPINN displays noticeable time offsets. The full-time error metric in Fig. 8b quantifies this gap: QCPIKAN reaches $1.3342\times10^{-2}$ versus QCPINN's $5.0634\times10^{-2}$ (3.8 times larger), proving QCPIKAN more accurately characterizes concentration transport and overall concentration field evolution.

Fig. 9 compares temporal concentration gradients and their errors. Gradients reflect solute front velocity and local variation magnitude, imposing stricter physical consistency demands. Fig. 9a shows both models capture major gradient peak locations and trends, but QCPIKAN's gradient curve aligns better with the benchmark at peaks and decaying segments, while QCPINN suffers severe local oscillations and deviations. As shown in Fig. 9b, their gradient errors are $5.5755\times10^{-2}$ and $1.6244\times10^{-1}$ respectively. QCPINN's error triples that of QCPIKAN. This demonstrates QCPIKAN superiorly resolves sharp local variations, front propagation velocity and transient dynamic responses.

Table 2. Model-configuration parameters of QCPINN and QCPIKAN in example 2.*2*

| Configuration | QCPINN | QCPIKAN |
|---|---|---|
| Quantum-circuit topology | Cascade | Cascade |
| Number of qubits | 6 | 6 |
| Quantum layers | 1 | 1 |
| Classical-network architecture | [3, 9, 1] | [3, 30, 1] |

| Trainable parameters | 565 | 594 |
|---|---|---|
| Learning rate | 0.001 | 0.001 |
| Batch size | 256 | 256 |

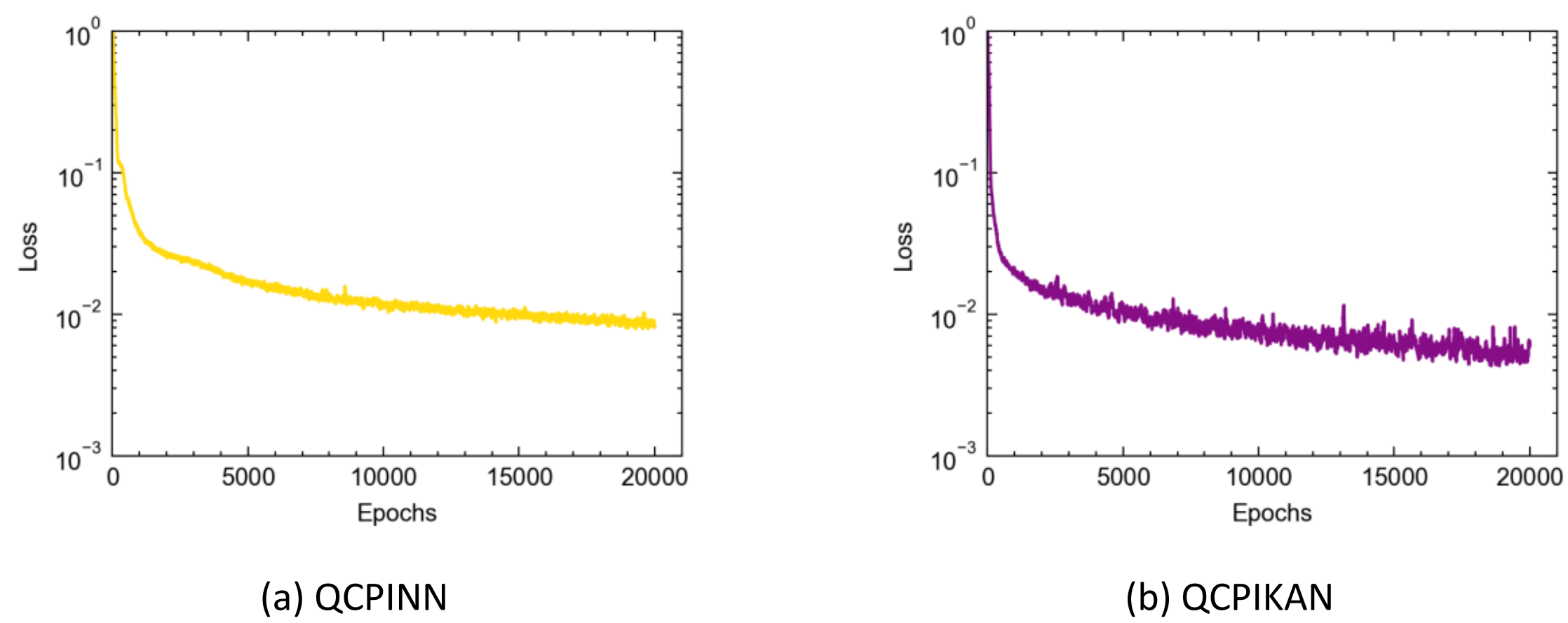

(a) QCPINN (b) QCPIKAN

Fig. 5. Training-loss curves of different methods in example 2.

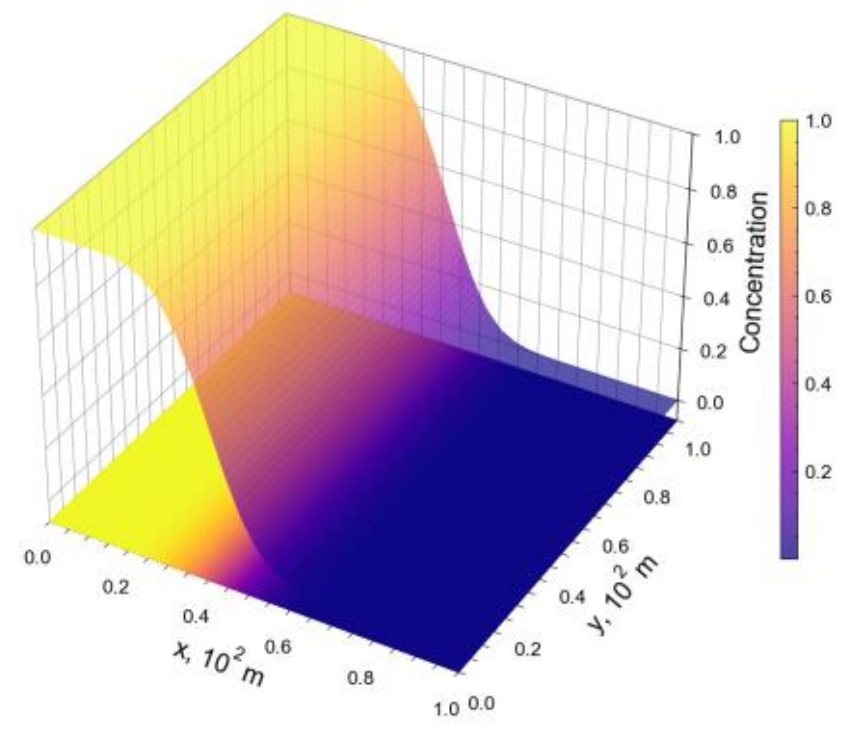

(a) Reference solution, $t$ = 0.4

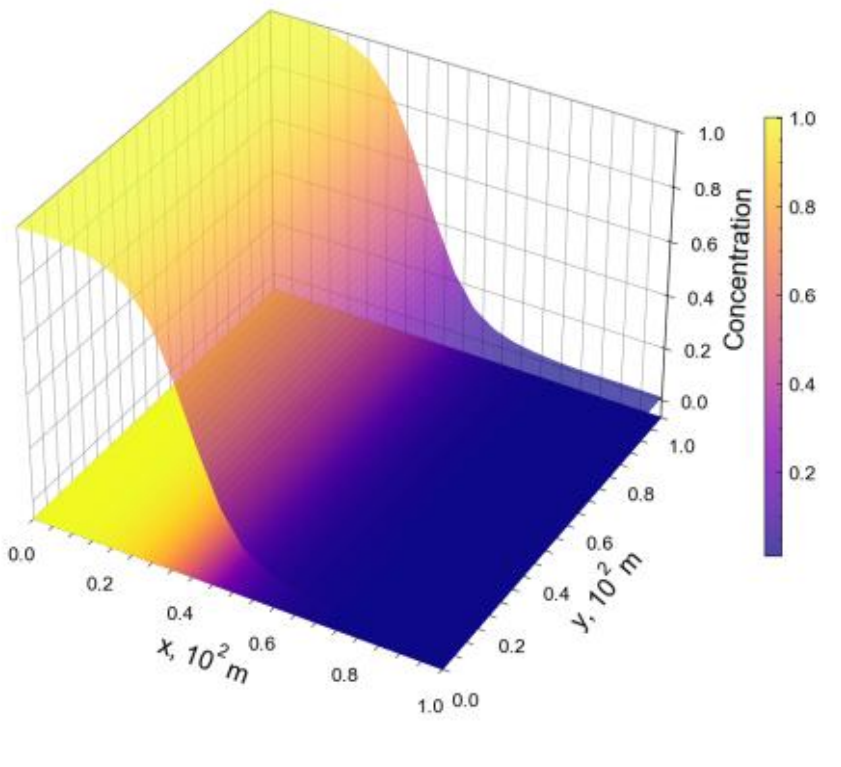

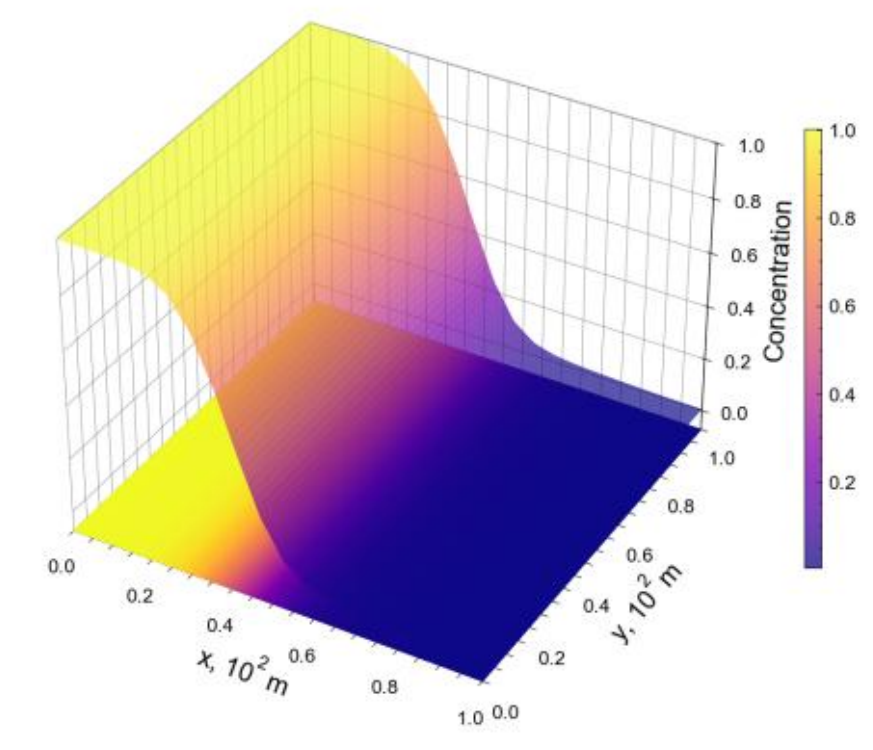

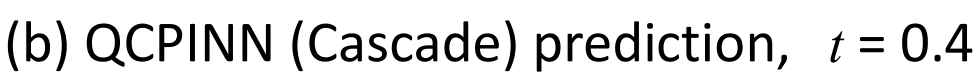

(b) QCPINN (Cascade) prediction, $t = 0.4$ (c) QCPIKAN (Cascade) prediction, $t = 0.4$

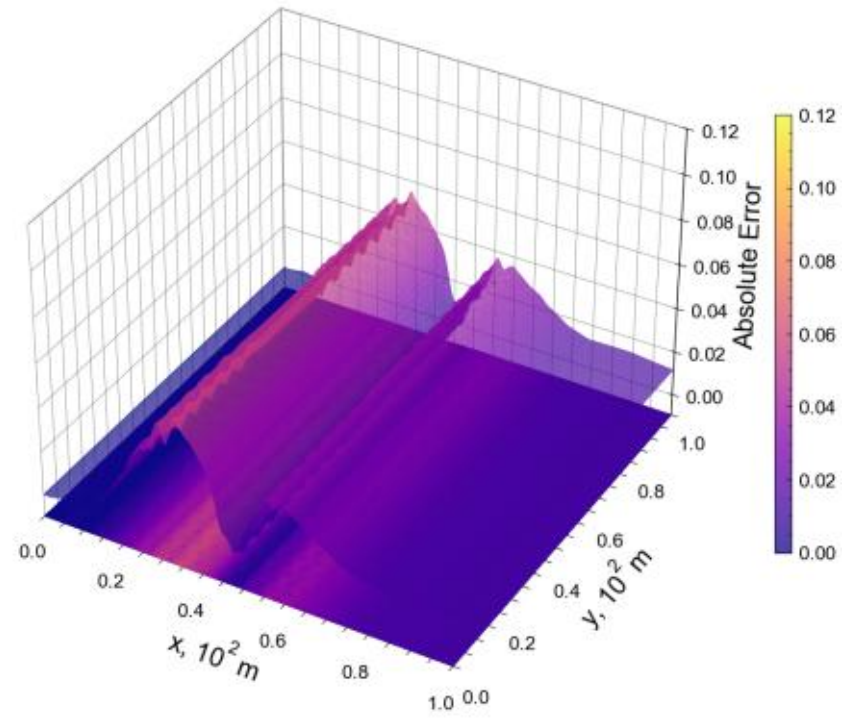

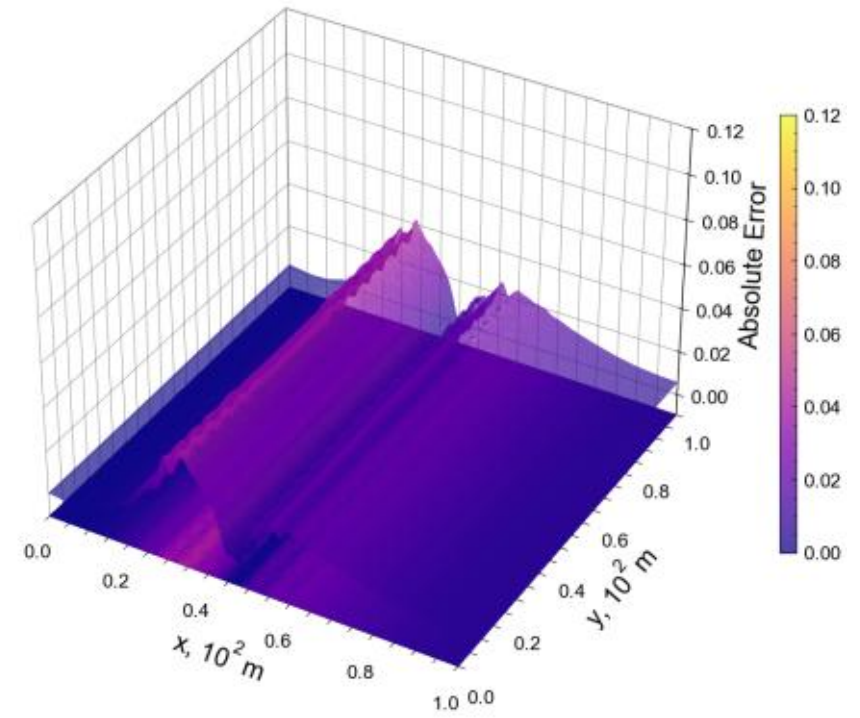

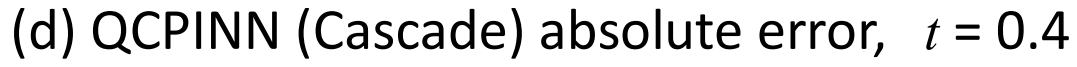

(d) QCPINN (Cascade) absolute error, $t = 0.4$ (e) QCPIKAN (Cascade) absolute error, $t = 0.4$

Fig. 6. Concentration distributions predicted by different methods and the corresponding error distributions at $t = 0.4$ in example 2.

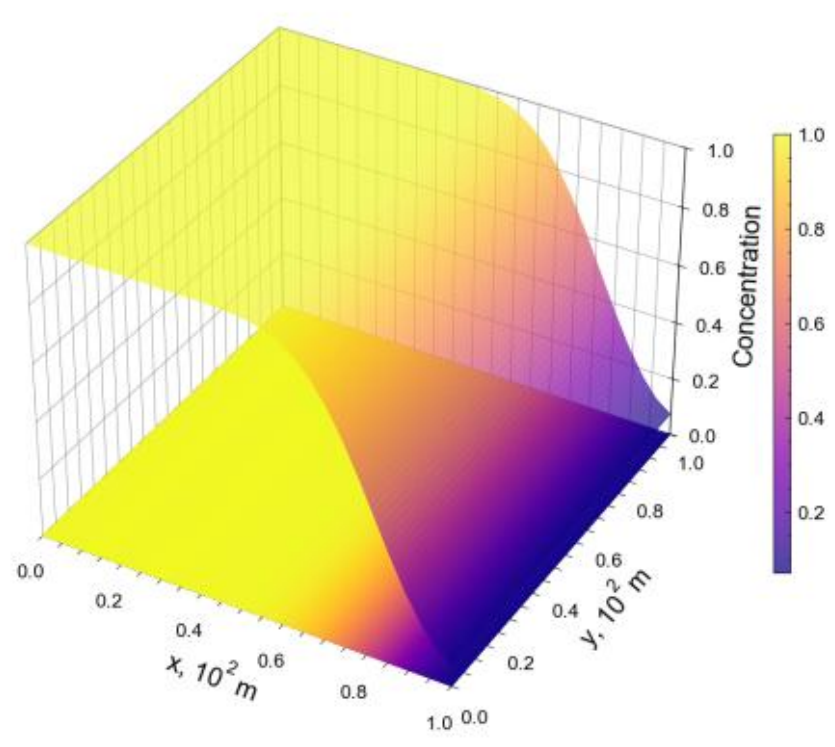

(a) Reference solution, $t = 0.8$

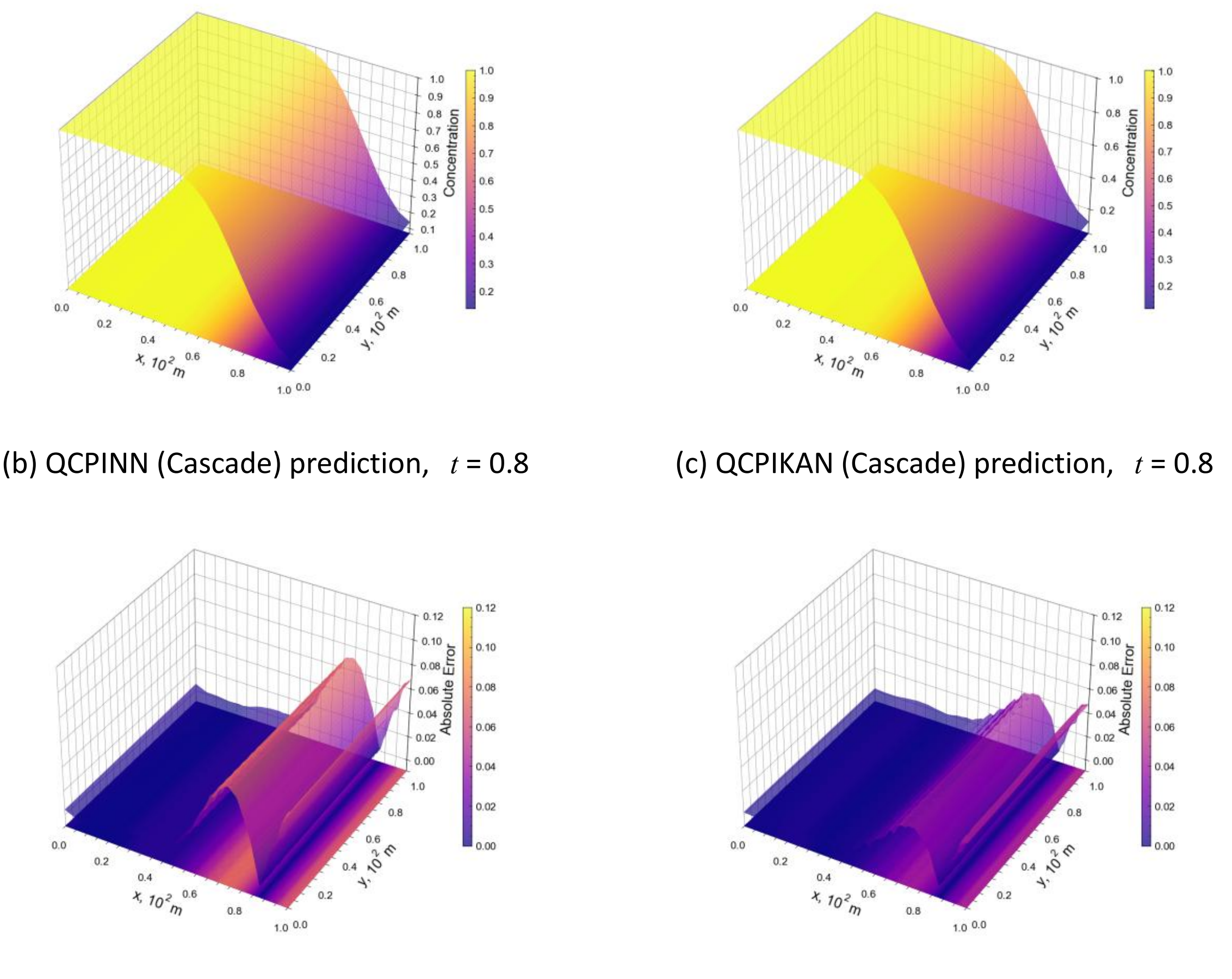

(b) QCPINN (Cascade) prediction, $t$ = 0.8

(c) QCPIKAN (Cascade) prediction, $t$ = 0.8

(d) QCPINN (Cascade) absolute error, $t$ = 0.8

(e) QCPIKAN (Cascade) absolute error, $t$ = 0.8

Fig. 7. Concentration distributions predicted by different methods and the corresponding error distributions at $t$ = 0.8 in example 2.

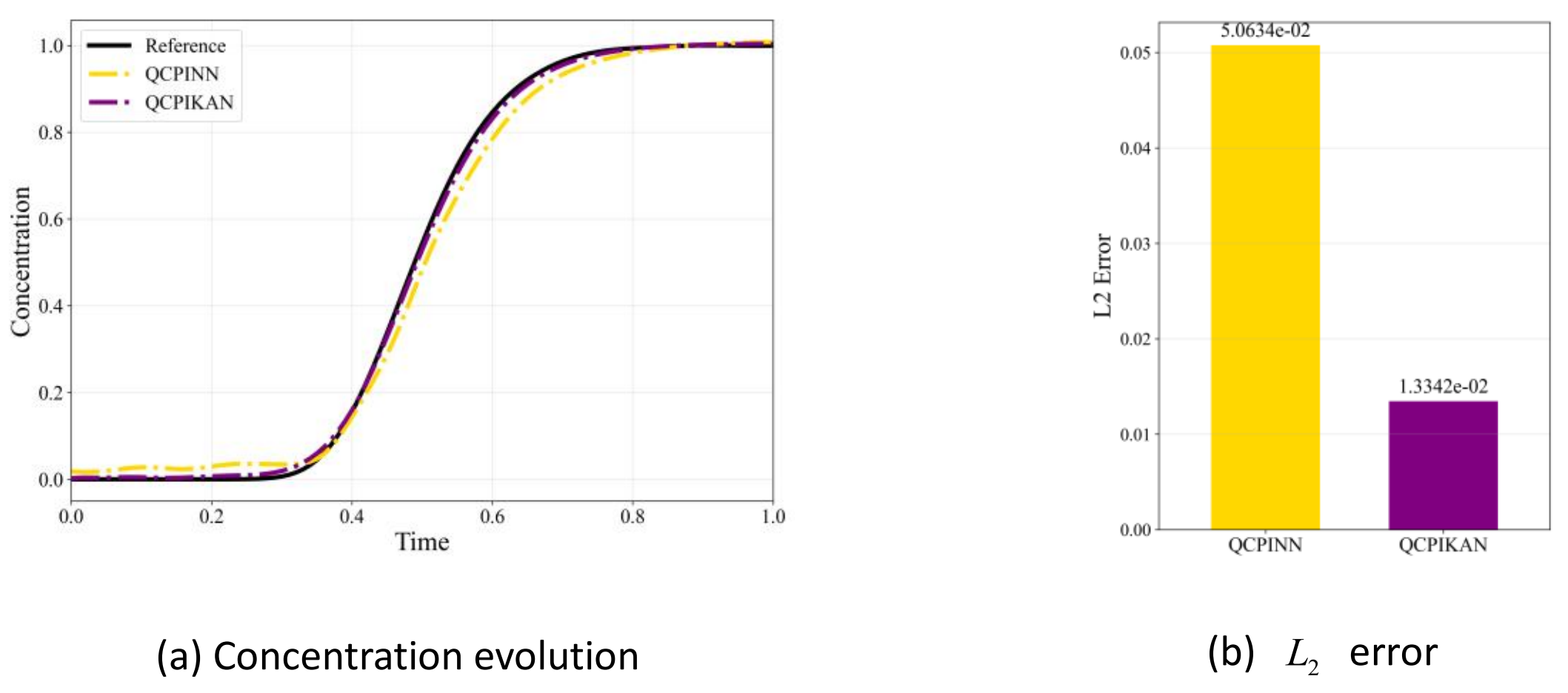

(a) Concentration evolution

(b) $L_2$ error

Fig. 8. Concentration prediction at the fixed point (50, 50) in example 2.

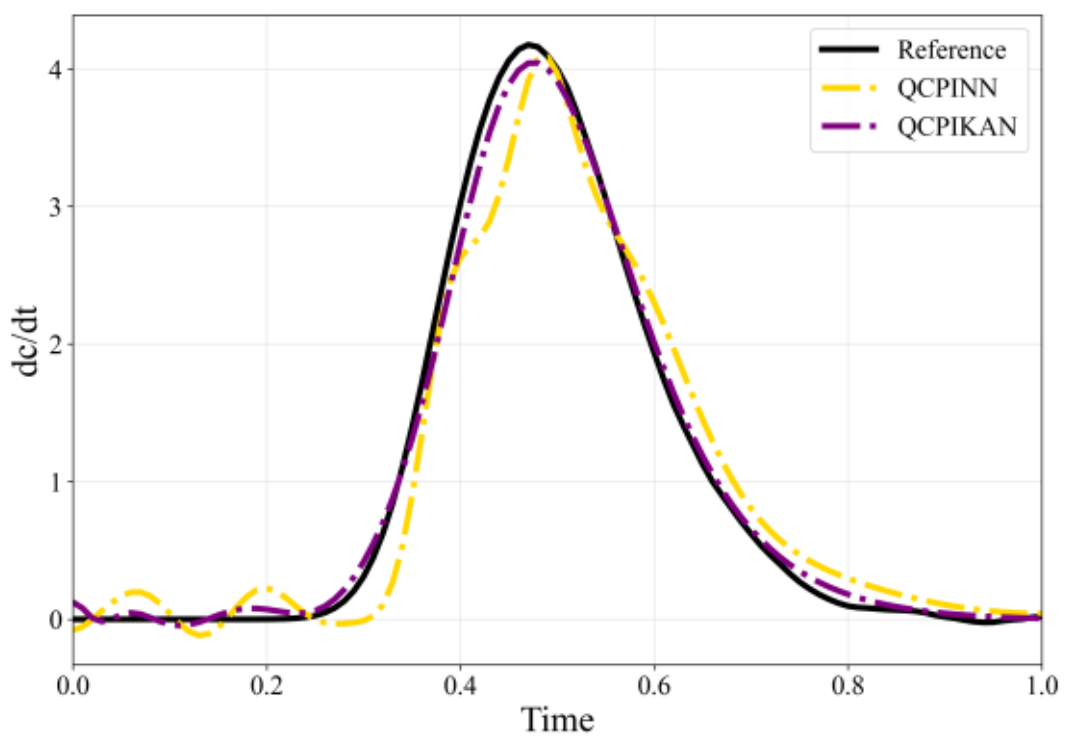

(a) Concentration-gradient evolution

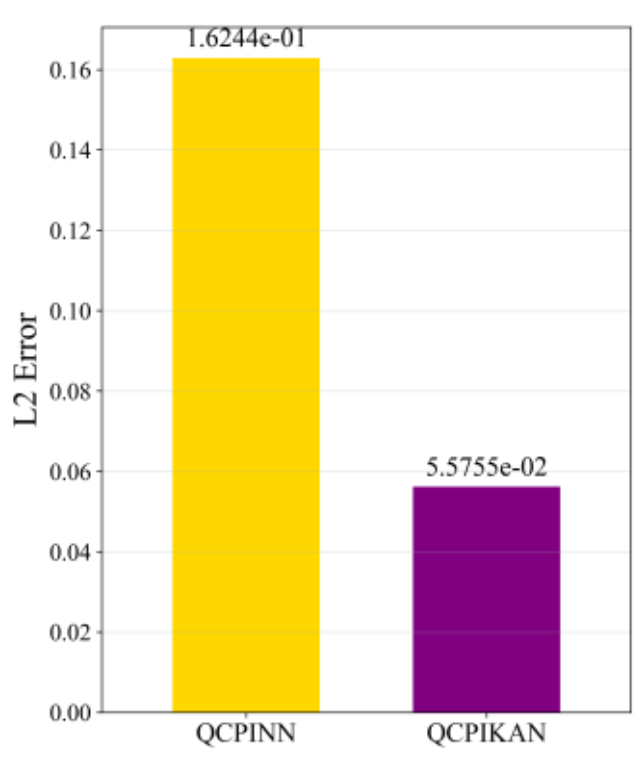

(b) $L_2$ error

Fig. 9. Concentration-gradient prediction at the fixed point (50, 50) in example 2.

*Example 3: transient oil-water two-phase seepage*

When capillary pressure, gravity and source/sink terms are neglected, two-phase incompressible seepage is described by the coupled mass-conservation equations of the water and oil phases:

$$\phi \frac{\partial S_w}{\partial t} + \nabla \cdot \left[ -\frac{k(x) k_{rw}(S_w)}{\mu_w} \nabla p \right] = 0, \tag{29}$$

$$\phi \frac{\partial (1 - S_w)}{\partial t} + \nabla \cdot \left[ -\frac{k(x) k_{ro}(S_w)}{\mu_o} \nabla p \right] = 0, \tag{30}$$

where the unknown functions to be solved are oil-phase pressure $p(x,t)$ and water saturation $S_w(x,t)$, both of which evolve in a coupled manner within the spatiotemporal domain $\Omega \times [0,T]$. $\phi$ denotes the rock porosity; $k(x)$ represents the absolute permeability of the rock; $\mu_w$ and $\mu_o$ are the water and oil phases, respectively; $k_{rw}(S_w)$ and $k_{ro}(S_w)$ are relative permeabilities of the water and oil phases, respectively. These are generally nonlinear functions of water saturation and are typically expressed using the Corey model:

$$k_{rw}(S_w) = \left( \frac{S_w - S_{wc}}{1 - S_{wc} - S_{or}} \right)^{n_w}, \quad k_{ro}(S_w) = \left( \frac{1 - S_w - S_{or}}{1 - S_{wc} - S_{or}} \right)^{n_o}. \tag{31}$$

The coupled equations formed by Eq. (29) and Eq. (30) contain both diffusive and convective characteristics. The left boundary is the injection boundary, maintaining a prescribed pressure of $p = 10\text{MPa}$ and water saturation of $S_w = 1$; the right boundary is the production boundary with $p = 5\text{MPa}$; and the top and bottom boundaries are closed boundaries, satisfying $\partial p / \partial y = 0$ and $\partial S_w / \partial y = 0$, implying no fluid inflow or outflow. The permeability adopts an exponential heterogeneous distribution in Eq. (32).

$$k(x,y) = 0.01 \times \exp\left(\frac{x^2+y^2}{5000}\right) D. \tag{32}$$

The porosity is set to $\phi = 0.2$, and the fluid viscosities are $\mu_w = 1\text{cP}$ and $\mu_o = 1\text{cP}$. In the Corey model, $S_{wc} = 0$, $S_{or} = 0$, $n_w = 1$, $n_o = 1$. Thus, we have $k_{rw}(S_w) = S_w$ and $k_{ro}(S_w) = 1 - S_w$. Under these parameters, the total mobility is $\lambda_t \equiv 1$, and the water cut is $f_w(S_w) = S_w$. The initial conditions are set as global $S_w = 0$ and $p = 5\text{MPa}$ .The network input is the spatiotemporal coordinate ( $t, x, y$ ), and the output is the pressure $p$ and water saturation $S_w$.

Table 3 lists the different hyperparameters of QCPINN and QCPIKAN, both using dual networks to approximate pressure and saturation individually. Fig. 10 shows their training loss curves. High-density high-order finite difference results serve as reference data. The pressure field in Figs. 11 and 12 is spatially smooth with low-frequency continuous characteristics. At *t*=0.33, QCPINN's maximum absolute error exceeds 0.096, while QCPIKAN's error stays near zero, reducing peak error by over an order of magnitude. At *t*=0.49, QCPINN's maximum error remains above 0.049 with clustered local errors; QCPIKAN also has concentrated errors but far smaller high-error zones. QCPIKAN delivers more uniform, precise pressure predictions without severe extreme error accumulation.

For water saturation at *t*=0.33 in Figs. 13a-13c, the benchmark features an arched displacement front. QCPINN shifts the front and overwidens its transition zone, while QCPIKAN perfectly matches the front's location and shape. QCPINN's peak absolute error hits 0.700 along the front, versus a narrow high-error strip for QCPIKAN. At *t*=0.49, the reference front curvature weakens. QCPINN still predicts an excessively broad transition zone, whereas QCPIKAN's front matches the benchmark closely. QCPINN's peak error falls to 0.425 but spreads over a wider area, while QCPIKAN's errors are tightly confined near the front with much smaller magnitude and coverage, proving reliable long-term evolution forecasting.

Single-point temporal analysis at coordinate (80,80) quantifies water saturation prediction and sharp front capture capacity, as shown in Fig. 15a. The reference water saturation stays near zero before *t*≈0.5, then jumps abruptly to 1.0 as the front arrives. QCPINN's saturation rises prematurely at *t*≈0.3 with a transition band twice the true width; QCPIKAN aligns nearly perfectly, accurately reproducing front arrival and steep saturation jump. Corresponding error metrics in Fig. 15b read 0.25153 for QCPINN and merely 0.075765 for QCPIKAN, with the former over 3.3 times larger.

Water saturation rate curves in Fig. 16a characterize front transition intensity via sharp reference single peak at *t*≈0.52. QCPINN severely underestimates peak magnitude, while QCPIKAN's peak closely matches the benchmark's sharp profile and better captures abrupt front-driven variations. Its error (0.56265) is substantially lower than QCPINN's 0.91556 (over 1.6 times larger).

In summary, QCPINN performs adequately in smooth domains, yet QCPIKAN excels at resolving steep-gradient, highly nonlinear and nearly discontinuous local zones. The embedded KAN architecture strengthens representation of intricate seepage fronts, boosting local prediction accuracy and physical consistency for coupled complex flow simulations.

Table 3. Model-configuration parameters of QCPINN and QCPIKAN in example 3.*3*

| Configuration | QCPINN | QCPIKAN |
|---|---|---|

| Quantum-circuit topology | Cross-mesh | Cross-mesh |
|---|---|---|
| Number of qubits | 6 | 6 |
| Quantum layers | 1 | 1 |
| Classical-network architecture | [3, 11, 1] | [3, 40, 1] |
| Trainable parameters | 565 | 594 |
| Learning rate | 0.001 | 0.001 |
| Batch size | 256 | 256 |

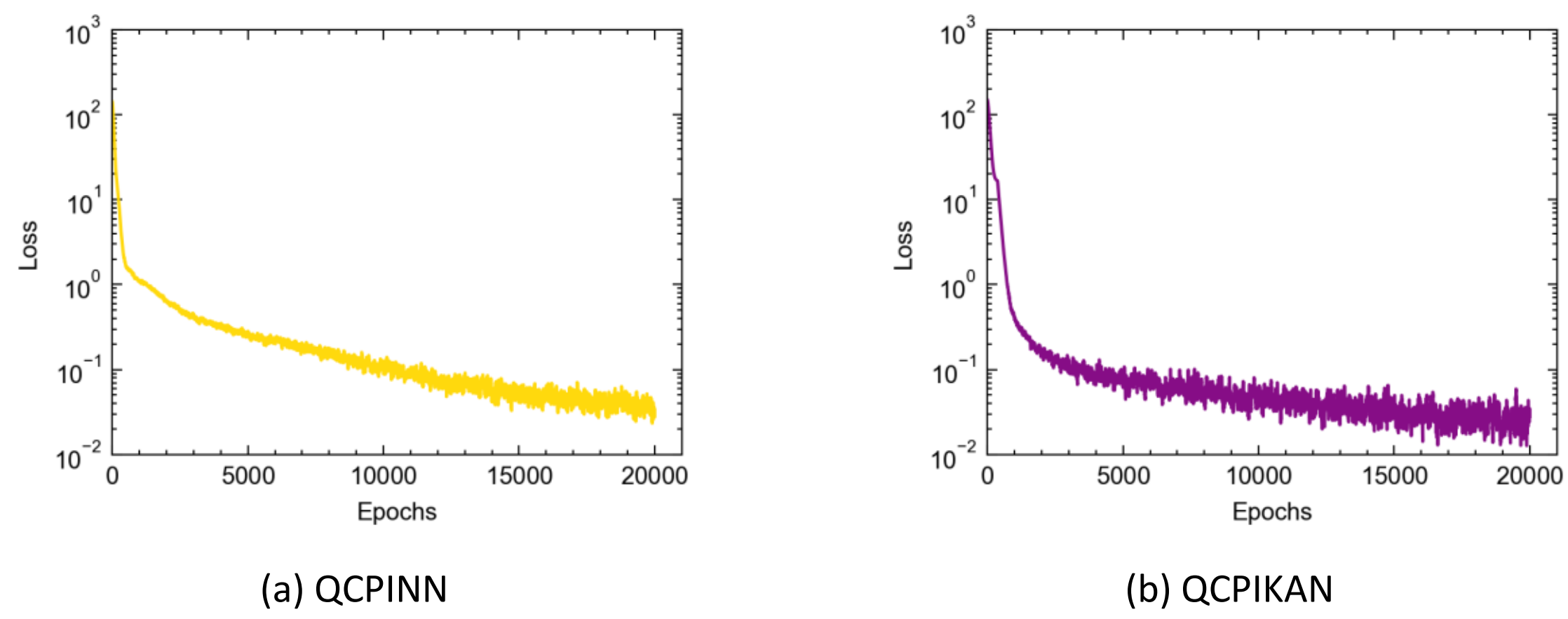

(a) QCPINN (b) QCPIKAN

Fig. 10. Training-loss curves of different methods in example 3.

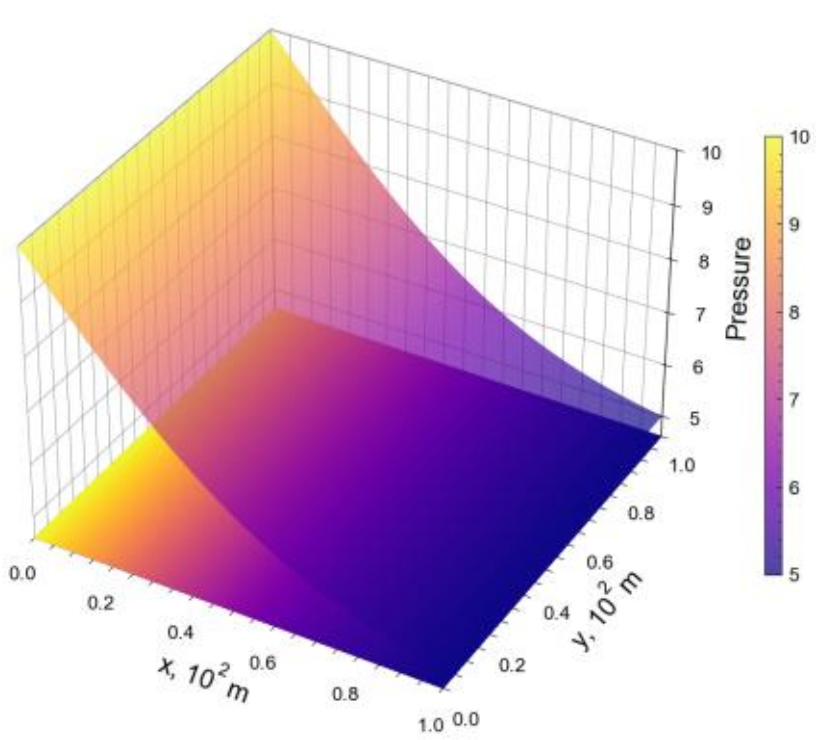

(a) Reference solution, $t$ = 0.33

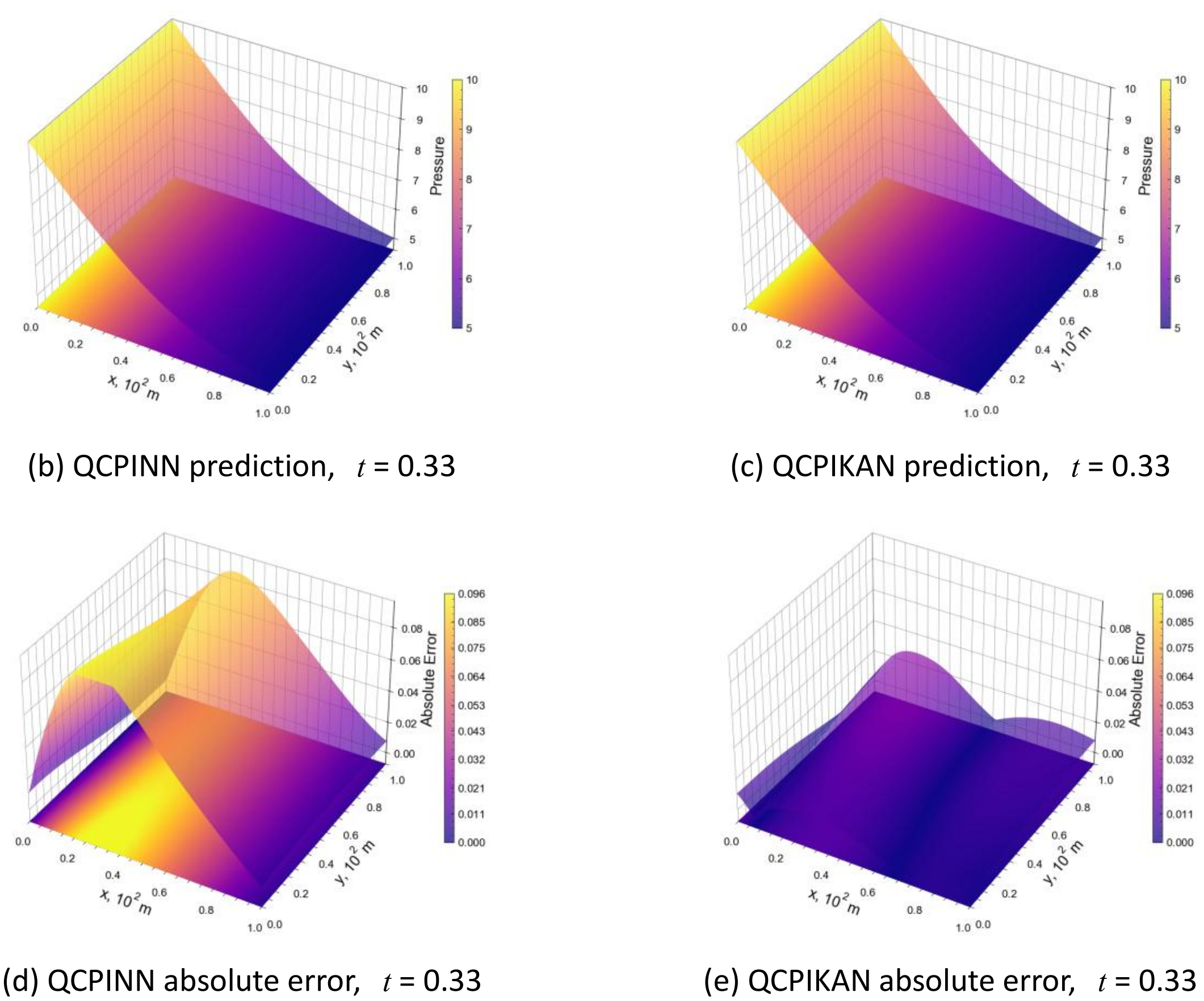

(b) QCPINN prediction, $t$ = 0.33

(c) QCPIKAN prediction, $t$ = 0.33

(d) QCPINN absolute error, $t$ = 0.33

(e) QCPIKAN absolute error, $t$ = 0.33

Fig. 11. Pressure distributions predicted by different methods and the corresponding error distributions at t = 0.33 in example 3.

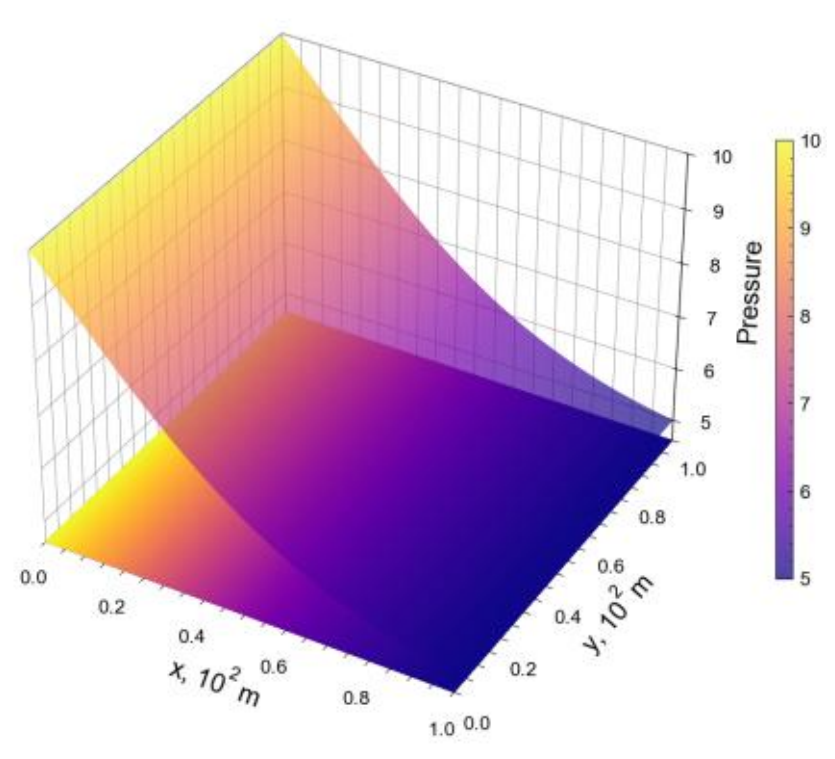

(a) Reference solution, $t$ = 0.49

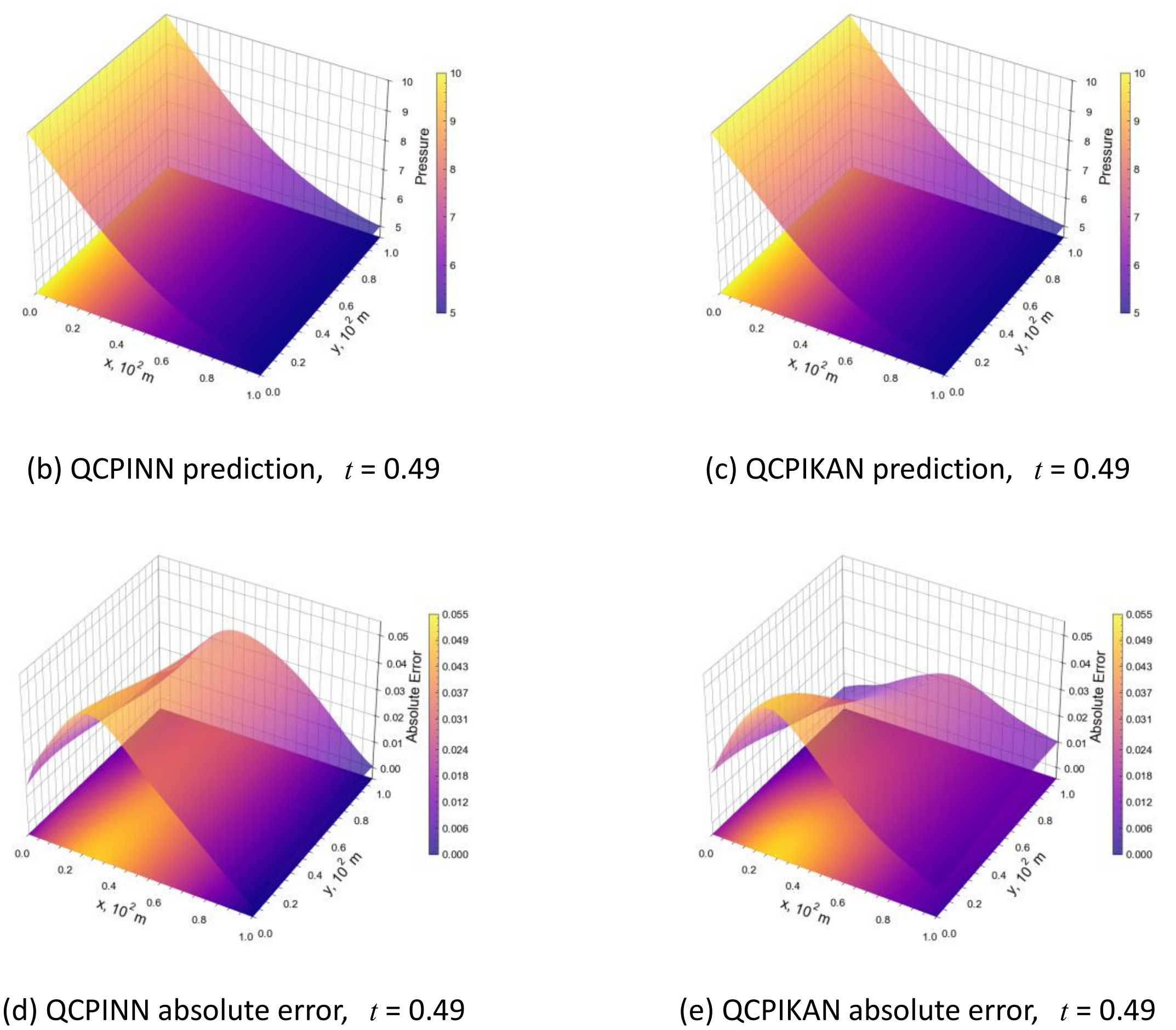

(b) QCPINN prediction, $t$ = 0.49

(c) QCPIKAN prediction, $t$ = 0.49

(d) QCPINN absolute error, $t$ = 0.49

(e) QCPIKAN absolute error, $t$ = 0.49

Fig. 12. Pressure distributions predicted by different methods and the corresponding error distributions at t = 0.49 in example 3.

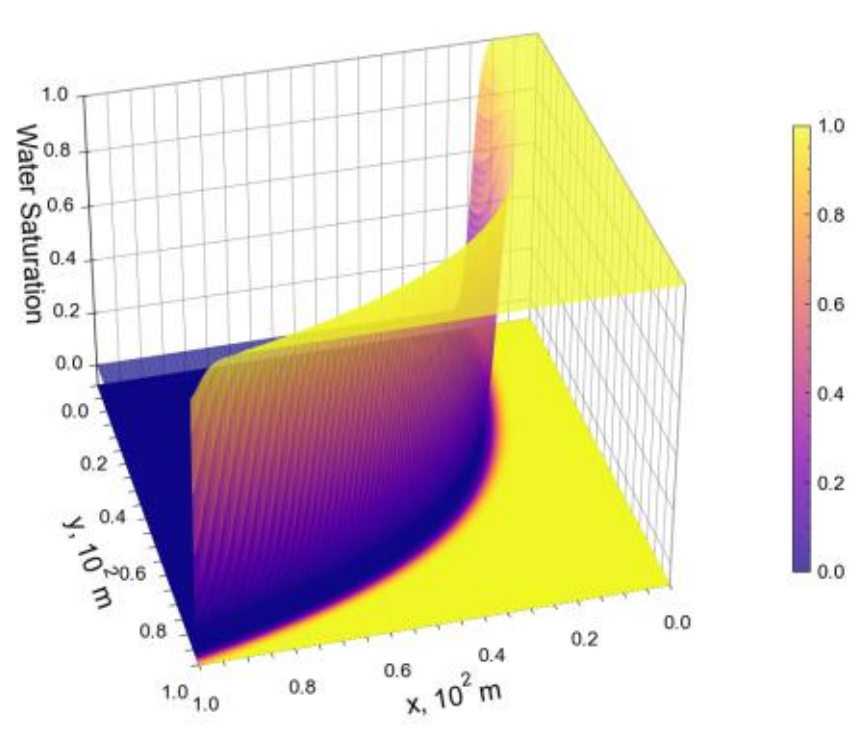

(a) Reference solution, $t$ = 0.33

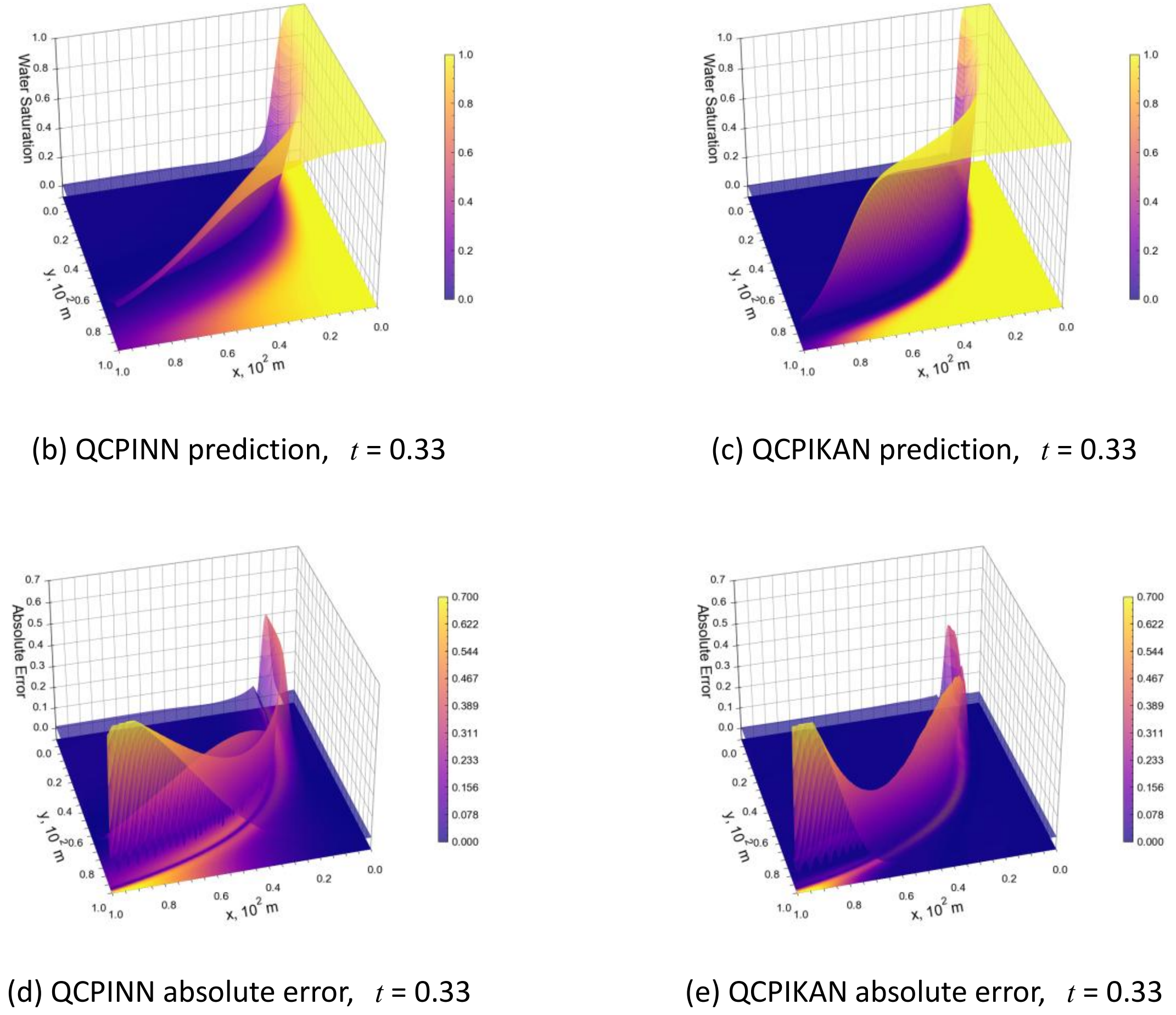

(b) QCPINN prediction, $t$ = 0.33

(c) QCPIKAN prediction, $t$ = 0.33

(d) QCPINN absolute error, $t$ = 0.33

(e) QCPIKAN absolute error, $t$ = 0.33

Fig. 13. Water-saturation distributions predicted by different methods and the corresponding error distributions at $t$ = 0.33 in example 3.

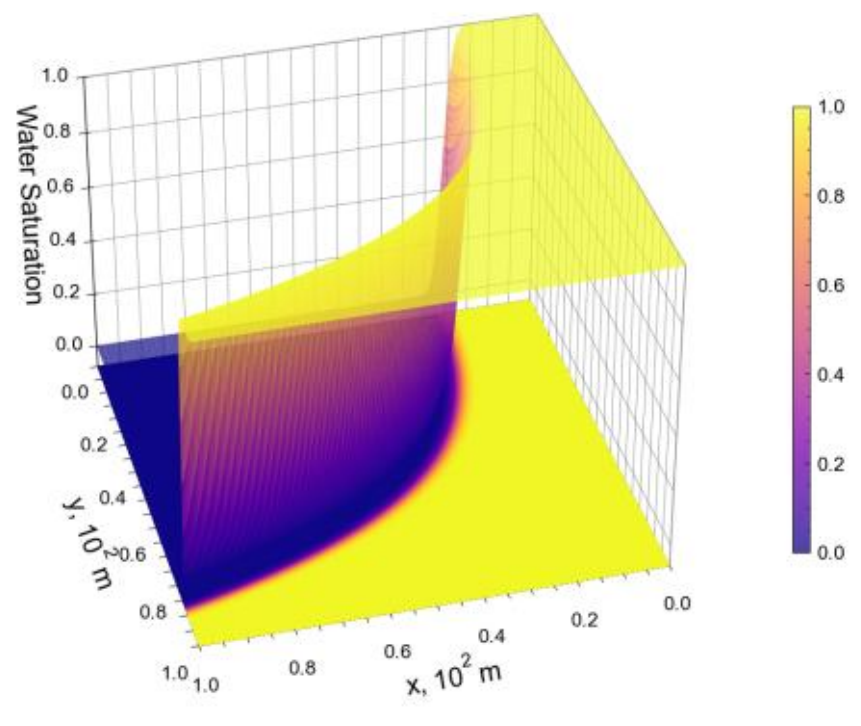

(a) Reference solution, $t$ = 0.49

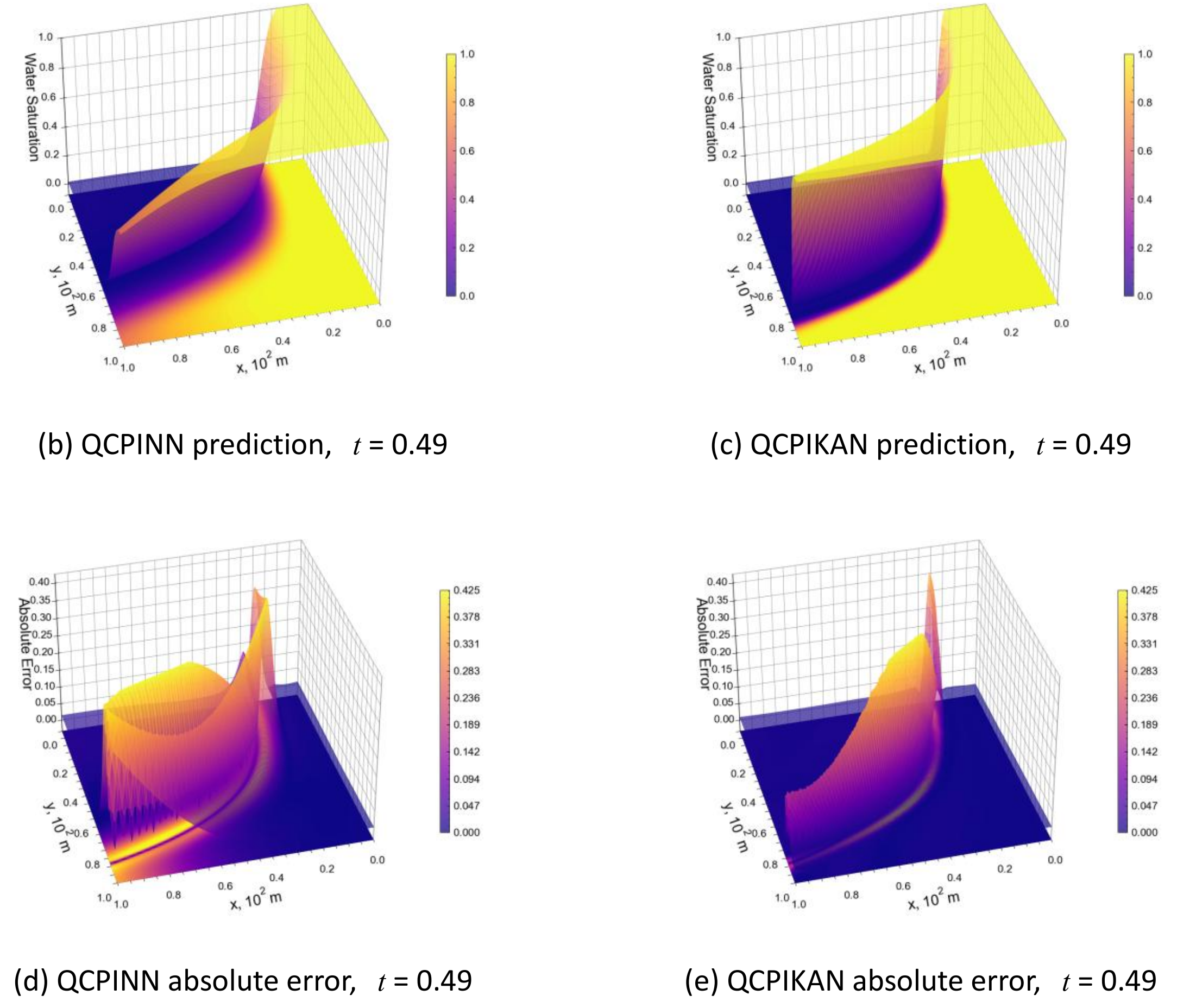

(b) QCPINN prediction, $t$ = 0.49

(c) QCPIKAN prediction, $t$ = 0.49

(d) QCPINN absolute error, $t$ = 0.49

(e) QCPIKAN absolute error, $t$ = 0.49

Fig. 14. Water-saturation distributions predicted by different methods and the corresponding error distributions at $t$ = 0.49 in example 3.

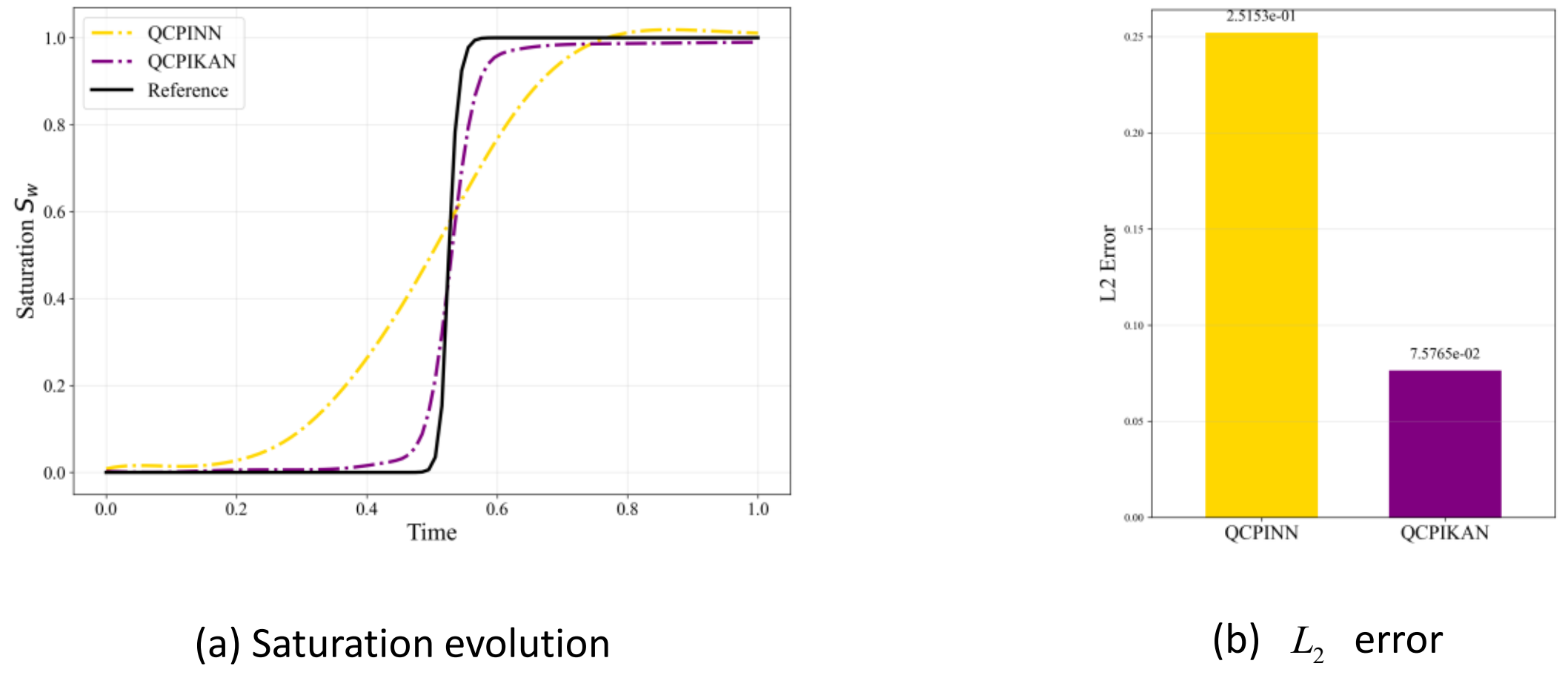

(a) Saturation evolution

(b) $L_2$ error

Fig. 15. Water-saturation prediction at the fixed point (80, 80) in example 3.

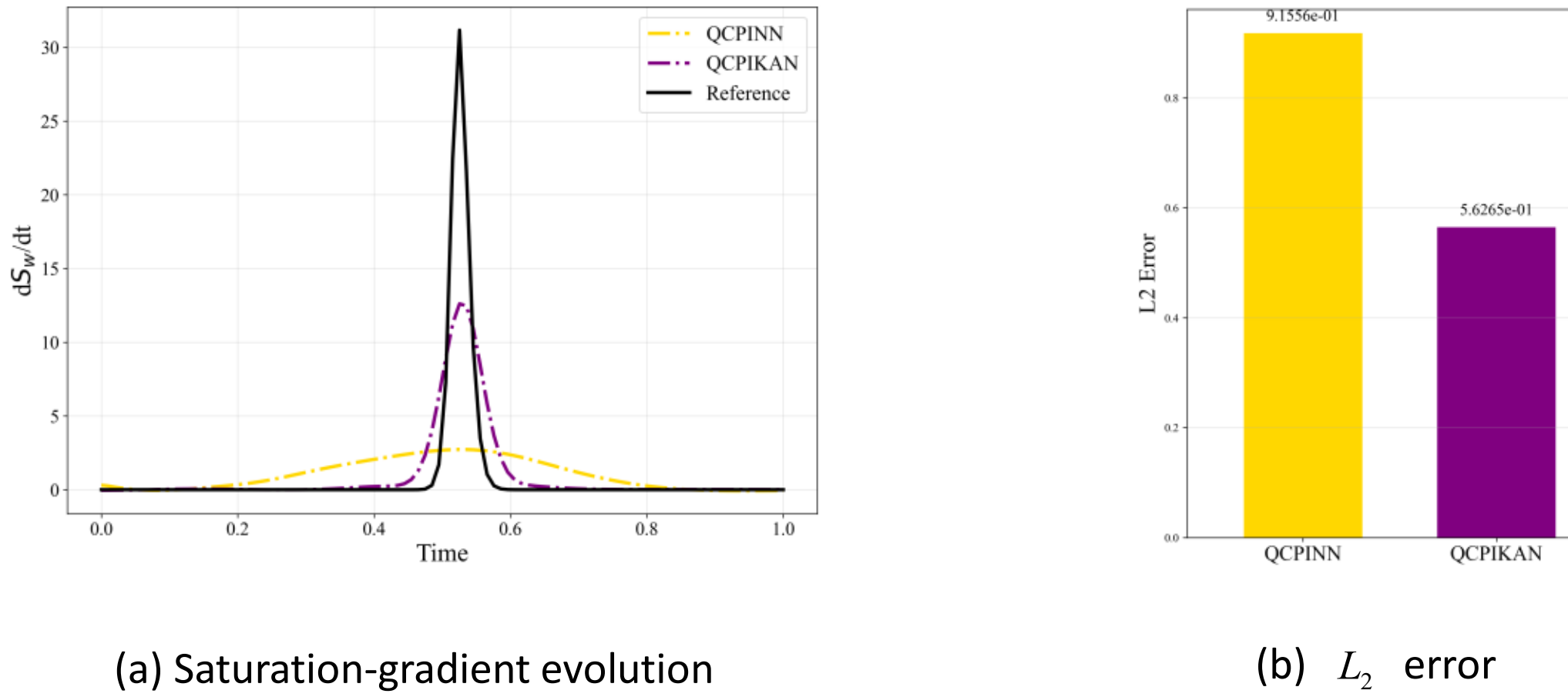

(a) Saturation-gradient evolution (b) $L_2$ error

Fig. 16. Saturation-gradient prediction at the fixed point (80, 80) in example 3.

## Discussion

We construct and validate QCPIKAN, the first hybrid quantum-classical physics-informed Kolmogorov-Arnold network. Our core architectural design substitutes standard MLP modules with Chebyshev-polynomial KAN layers for classical preprocessing and postprocessing, unlocking adaptive edge-function fitting and enhanced fine-grained nonlinear representation. Grounded in approximation theory, rigorous mathematical derivations demonstrate that the inherent spectral approximation characteristics of QCPIKAN lift high-frequency error convergence from algebraic to exponential order, fundamentally mitigating numerical dispersion within regions featuring steep gradients.

We systematically validate this theoretical advantage across three progressively complex PDE benchmarks representative of canonical physical solving scenarios: heterogeneous steady-state single-phase transport, transient advection-diffusion-adsorption evolution, and dynamically coupled two-variable physical systems. Quantitatively, QCPIKAN consistently outperforms conventional QCPINN across all numerical evaluations. For transient transport test cases, the concentration $L_2$ errors delivered by QCPINN are approximately 3.8 times greater, while its concentration-gradient errors are roughly 3 times larger relative to QCPIKAN. Within coupled two-variable physical simulations, QCPIKAN suppresses the maximum absolute pressure error by over one order of magnitude. Meanwhile, the saturation-associated $L_2$ and gradient errors of QCPINN reach approximately 3.3 times and 1.6 times the error magnitudes of QCPIKAN, respectively. Collectively, our theoretical derivations and numerical benchmarks confirm that QCPIKAN constitutes an accurate, physically consistent, and generalizable numerical framework for arbitrary complex partial differential equations. This work further establishes a versatile blueprint for hybrid quantum-classical scientific computation and expands the scope of spectral Kolmogorov-Arnold network architectures for broad PDE-governed physical modelling tasks.

## Acknowledgements

Prof. Rao acknowledges financial support from the General Program of National Natural Science

Foundation of China (NSFC) (No. 52574028), the National Science and Technology Major Project of China (No. 2025ZD1401106), the General Program of Natural Science Foundation of Hubei Province (No. 2026AFB760), and the Xinjiang Uygur Autonomous Region Scientific and Technological Innovation Team Project (No. 2024TSYCTD0018).

## References

[1] Raissi, Perdikaris, and Karniadakis, 'Physics-informed neural networks: A deep learning framework for solving forward and inverse problems involving nonlinear partial differential equations', 2019.

[2] G. E. Karniadakis, I. G. Kevrekidis, L. Lu, P. Perdikaris, S. Wang, and L. Yang, 'Physics-informed machine learning', *Nat Rev Phys*, vol. 3, no. 6, pp. 422–440, May 2021, doi: 10.1038/s42254-021-00314-5.

[3] Y. Yang, X. Wang, J. Li, and R. Ge, 'A data-physic driven method for gear fault diagnosis using PINN and pseudo-dynamic features', *Measurement*, vol. 236, p. 115124, Aug. 2024, doi: 10.1016/j.measurement.2024.115124.

[4] N. Ahmadi Daryakenari, M. De Florio, K. Shukla, and G. E. Karniadakis, 'AI-aristotle: A physics-informed framework for systems biology gray-box identification', *PLoS Comput Biol*, vol. 20, no. 3, p. e1011916, Mar. 2024, doi: 10.1371/journal.pcbi.1011916.

[5] J. Hua, Y. Li, C. Liu, P. Wan, and X. Liu, 'Physics-informed neural networks with weighted losses by uncertainty evaluation for accurate and stable prediction of manufacturing systems', *IEEE Trans. Neural Netw. Learning Syst.*, vol. 35, no. 8, pp. 11064–11076, Aug. 2024, doi: 10.1109/TNNLS.2023.3247163.

[6] Y. Xia and Y. Meng, 'Physics-informed neural network (PINN) for solving frictional contact temperature and inversely evaluating relevant input parameters', *Lubricants*, vol. 12, no. 2, p. 62, Feb. 2024, doi: 10.3390/lubricants12020062.

[7] A. M. Tartakovsky, C. O. Marrero, P. Perdikaris, G. D. Tartakovsky, and D. Barajas-Solano, 'Physics-informed deep neural networks for learning parameters and constitutive relationships in subsurface flow problems', *Water Resources Research*, vol. 56, no. 5, p. e2019WR026731, May 2020, doi: 10.1029/2019WR026731.

[8] J. M. Hanna, J. V. Aguado, S. Comas-Cardona, R. Askri, and D. Borzacchiello, 'Residual-based adaptivity for two-phase flow simulation in porous media using physics-informed neural networks', *Computer Methods in Applied Mechanics and Engineering*, vol. 396, p. 115100, Jun. 2022, doi: 10.1016/j.cma.2022.115100.

[9] O. Fuks and H. A. Tchelepi, 'Limitations of physics informed machine learning for nonlinear two-phase transport in porous media', *J Mach Learn Model Comput*, vol. 1, no. 1, pp. 19–37, 2020, doi: 10.1615/JMachLearnModelComput.2020033905.

[10] R. Xu, D. Zhang, M. Rong, and N. Wang, 'Weak form theory-guided neural network (TgNN-wf) for deep learning of subsurface single- and two-phase flow', *Journal of Computational Physics*, vol. 436, p. 110318, Jul. 2021, doi: 10.1016/j.jcp.2021.110318.

[11] M. M. Almajid and M. O. Abu-Al-Saud, 'Prediction of porous media fluid flow using physics informed neural networks', *Journal of Petroleum Science and Engineering*, vol. 208, p. 109205, Jan. 2022, doi: 10.1016/j.petrol.2021.109205.

[12] F. Lehmann, M. Fahs, A. Alhubail, and H. Hoteit, ‘A mixed pressure-velocity formulation to model flow in heterogeneous porous media with physics-informed neural networks’, *Advances in Water Resources*, vol. 181, p. 104564, Nov. 2023, doi: 10.1016/j.advwatres.2023.104564.

[13] Z. Fang, ‘A high-efficient hybrid physics-informed neural networks based on convolutional neural network’, *IEEE Trans. Neural Netw. Learning Syst.*, vol. 33, no. 10, pp. 5514–5526, Oct. 2022, doi: 10.1109/TNNLS.2021.3070878.

[14] X. Yan *et al.*, ‘Physics-informed neural network simulation of two-phase flow in heterogeneous and fractured porous media’, *Advances in Water Resources*, vol. 189, p. 104731, Jul. 2024, doi: 10.1016/j.advwatres.2024.104731.

[15] J.-Q. Lin *et al.*, ‘A layer-specific constraint-based enriched physics-informed neural network for solving two-phase flow problems in heterogeneous porous media’, *Petroleum Science*, vol. 22, no. 11, pp. 4714–4735, Nov. 2025, doi: 10.1016/j.petsci.2025.07.008.

[16] Z. Liu *et al.*, ‘KAN: Kolmogorov-arnold networks’, Feb. 09, 2025, *arXiv*: arXiv:2404.19756. doi: 10.48550/arXiv.2404.19756.

[17] K. Shukla, J. D. Toscano, Z. Wang, Z. Zou, and G. E. Karniadakis, ‘A comprehensive and FAIR comparison between MLP and KAN representations for differential equations and operator networks’, *Computer Methods in Applied Mechanics and Engineering*, vol. 431, p. 117290, Nov. 2024, doi: 10.1016/j.cma.2024.117290.

[18] C. Zeng, J. Wang, H. Shen, and Q. Wang, ‘KAN versus MLP on irregular or noisy function’, in *2025 15th IEEE International Conference on Pattern Recognition Systems (ICPRS)*, Valparaiso,Viña del Mar, Chile: IEEE, Dec. 2025, pp. 1–7. doi: 10.1109/ICPRS66293.2025.11302852.

[19] H. Shuai and F. Li, ‘Physics-informed kolmogorov-arnold networks for power system dynamics’, *IEEE Open J. Power Energy*, vol. 12, pp. 46–58, 2025, doi: 10.1109/OAJPE.2025.3529928.

[20] B. Jiang, Y. Wang, Q. Wang, and H. Geng, ‘A hybrid KAN-ANN based model for interpretable and enhanced short-term load forecasting’, *IEEE Trans. Artif. Intell.*, pp. 1–12, 2026, doi: 10.1109/TAI.2026.3656248.

[21] S. Lin *et al.*, ‘KAN-UG: Kolmogorov–arnold network-based difference feature enhancement method with uncertainty guidance for change detection’, *IEEE Trans. Geosci. Remote Sensing*, vol. 64, pp. 1–15, 2026, doi: 10.1109/TGRS.2026.3681936.

[22] M.-Z. Huang, Y.-X. Peng, Z.-T. Jiang, P.-N. Sun, and C.-X. Jiang, ‘Physics-informed KAN-coupled FEM for deformation analysis of complex shells’, *Thin-Walled Structures*, vol. 216, p. 113725, Nov. 2025, doi: 10.1016/j.tws.2025.113725.

[23] Z.-T. Jiang, M.-Z. Huang, Y.-X. Peng, C.-X. Jiang, and W.-W. Wu, ‘Engineering application and experimental validation of the FEM-PIKAN model in ship hull deformation analysis’, *Ocean Engineering*, vol. 344, p. 123588, Jan. 2026, doi: 10.1016/j.oceaneng.2025.123588.

[24] Y. Gong, Y. He, Y. Mei, X. Zhuang, F. Qin, and T. Rabczuk, ‘Physics-informed kolmogorov-arnold networks for multi-material elasticity problems in electronic packaging’, *Applied Mathematical Modelling*, vol. 156, p. 116793, Aug. 2026, doi: 10.1016/j.apm.2026.116793.

[25] J. Wan *et al.*, 'PIKANs: Physics-informed kolmogorov–arnold networks for landslide time-to-failure prediction', *Computers & Geosciences*, vol. 208, p. 106094, Feb. 2026, doi: 10.1016/j.cageo.2025.106094.

[26] X. Rao, Y. Liu, X. He, and H. Hoteit, 'Physics-informed kolmogorov–arnold networks to model flow in heterogeneous porous media with a mixed pressure-velocity formulation', *Physics of Fluids*, vol. 37, no. 7, p. 076654, Jul. 2025, doi: 10.1063/5.0279122.

[27] T. Imankulov, Y. Kenzhebek, S. D. Bekele, and D. Akhmed-Zaki, 'Performance and stability of physics-informed kolmogorov-arnold networks for two-phase transport in porous media: A comparative study with physics-informed neural networks', *Results in Engineering*, vol. 29, p. 108823, Mar. 2026, doi: 10.1016/j.rineng.2025.108823.

[28] Q. Wang, W. Zha, D. Li, X. Li, L. Shen, and Z. Shi, 'Parameterized analytical solution of seepage equation for reservoir simulation using physics-informed kolmogorov-arnold network without labels', *Journal of Computational Physics*, vol. 548, p. 114599, Mar. 2026, doi: 10.1016/j.jcp.2025.114599.

[29] J. Preskill, 'Quantum computing in the NISQ era and beyond', *Quantum*, vol. 2, p. 79, Aug. 2018, doi: 10.22331/q-2018-08-06-79.

[30] K. Bharti *et al.*, 'Noisy intermediate-scale quantum algorithms', *Rev. Mod. Phys.*, vol. 94, no. 1, p. 015004, Feb. 2022, doi: 10.1103/RevModPhys.94.015004.

[31] E. A. R. Guzman and D. Lacroix, 'Restoring symmetries in quantum computing using classical shadows', Nov. 08, 2023, *arXiv*: arXiv:2311.04571. doi: 10.48550/arXiv.2311.04571.

[32] J. Villalba-Díez and J. Ordieres-Meré, 'Quantum-enhanced signal processing via VQE for improved biomechanical feedback control', *Digital Signal Processing*, vol. 166, p. 105357, Nov. 2025, doi: 10.1016/j.dsp.2025.105357.

[33] A. Bayro and H. Jeong, 'Enhancing emotional response detection in virtual reality with quantum support vector machine learning', *Computers & Graphics*, vol. 128, p. 104196, May 2025, doi: 10.1016/j.cag.2025.104196.

[34] J. Choi *et al.*, 'Performance evaluation of quantum support vector machine for COVID-19 biomarker analysis', *Computer Methods and Programs in Biomedicine*, vol. 281, p. 109343, Jul. 2026, doi: 10.1016/j.cmpb.2026.109343.

[35] C. Bravo-Prieto, R. LaRose, M. Cerezo, Y. Subasi, L. Cincio, and P. J. Coles, 'Variational quantum linear solver', *Quantum*, vol. 7, p. 1188, Nov. 2023, doi: 10.22331/q-2023-11-22-1188.

[36] Y. Y. Liu *et al.*, 'Application of a variational hybrid quantum-classical algorithm to heat conduction equation and analysis of time complexity', *Physics of Fluids*, vol. 34, no. 11, p. 117121, Nov. 2022, doi: 10.1063/5.0121778.

[37] X. Rao, 'Performance study of variational quantum linear solver with an improved ansatz for reservoir flow equations', *Physics of Fluids*, vol. 36, no. 4, p. 047104, Apr. 2024, doi: 10.1063/5.0201739.

[38] M. G. Meena *et al.*, 'Assessing VQLS for fluid dynamics on a hybrid quantum-HPC stack', in *2025 IEEE International Conference on Quantum Computing and Engineering (QCE)*, Albuquerque, NM, USA: IEEE, Aug. 2025, pp. 484–485. doi: 10.1109/QCE65121.2025.10407.

[39] S. K. N. Kannan and S. Kamrava, 'Quantum computing for modeling fluid flow in subsurface

environments', *Advances in Water Resources*, vol. 215, p. 105383, Sep. 2026, doi: 10.1016/j.advwatres.2026.105383.

[40] X. Rao, 'The first application of quantum computing algorithm in streamline-based simulation of water-flooding reservoirs', in *ADIPEC*, Abu Dhabi, UAE: SPE, Nov. 2024, p. D011S003R006. doi: 10.2118/221850-MS.

[41] L. A. Moncayo-Martínez and N. He, 'Quantum optimisation for supply chain: QUBO formulations and QAOA solutions for facility location and load balancing', *Results in Engineering*, vol. 29, p. 108373, Mar. 2026, doi: 10.1016/j.rineng.2025.108373.

[42] R. Tate and S. Eidenbenz, 'Theoretical approximation ratios for warm-started QAOA on 3-regular max-cut instances at depth $p$ =', *Theoretical Computer Science*.

[43] V. Havlíček, 'Supervised learning with quantum-enhanced feature spaces'.

[44] L. Gonon and A. Jacquier, 'Universal approximation theorem and error bounds for quantum neural networks and quantum reservoirs', *IEEE Trans. Neural Netw. Learning Syst.*, vol. 36, no. 6, pp. 11355–11368, Jun. 2025, doi: 10.1109/TNNLS.2025.3552223.

[45] A. Abbas, D. Sutter, C. Zoufal, A. Lucchi, A. Figalli, and S. Woerner, 'The power of quantum neural networks', *Nat Comput Sci*, vol. 1, no. 6, pp. 403–409, Jun. 2021, doi: 10.1038/s43588-021-00084-1.

[46] M. Cerezo, A. Sone, T. Volkoff, L. Cincio, and P. J. Coles, 'Cost function dependent barren plateaus in shallow parametrized quantum circuits', *Nat Commun*, vol. 12, no. 1, p. 1791, Mar. 2021, doi: 10.1038/s41467-021-21728-w.

[47] R. Rahman and D. C. Nguyen, 'Quantum convolutional neural networks: A survey on architectures, applications, and future directions', *IEEE Trans. Neural Netw. Learning Syst.*, pp. 1–20, 2026, doi: 10.1109/TNNLS.2026.3677762.

[48] O. Kyriienko, A. E. Paine, and V. E. Elfving, 'Solving nonlinear differential equations with differentiable quantum circuits', *Phys. Rev. A*, vol. 103, no. 5, p. 052416, May 2021, doi: 10.1103/PhysRevA.103.052416.

[49] S. N. Larki, M. Mosleh, and M. Kheyrandish, 'Towards quantum audio steganalysis using synergy of quantum fourier transform and quantum neural network', *Engineering Applications of Artificial Intelligence*, vol. 159, p. 111595, Nov. 2025, doi: 10.1016/j.engappai.2025.111595.

[50] J. Tian *et al.*, 'Recent advances for quantum neural networks in generative learning', *IEEE Trans. Pattern Anal. Mach. Intell.*, vol. 45, no. 10, pp. 12321–12340, Oct. 2023, doi: 10.1109/TPAMI.2023.3272029.

[51] Y. Peng, X. Li, Z. Liang, and Y. Wang, 'HyQ2: A hybrid quantum neural network for NextG vulnerability detection', *IEEE Trans. Quantum Eng.*, vol. 5, pp. 1–19, 2024, doi: 10.1109/TQE.2024.3481280.

[52] J. Zheng, Q. Gao, M. Ogorzałek, J. Lü, and Y. Deng, 'A quantum spatial graph convolutional neural network model on quantum circuits', *IEEE Trans. Neural Netw. Learning Syst.*, vol. 36, no. 3, pp. 5706–5720, Mar. 2025, doi: 10.1109/TNNLS.2024.3382174.

[53] X. Rao, C. Luo, X. He, and K. Hyung, 'An efficient quantum neural network model for prediction of carbon dioxide CO2 sequestration in saline aquifers', in *ADIPEC*, Abu Dhabi, UAE: SPE, Nov. 2024, p. D021S061R005. doi: 10.2118/222257-MS.

[54] H. Cho, J. Kim, K. T. No, and H. Lim, ‘Hybrid quantum neural networks with variational quantum regressor for enhancing QSPR modeling of CO2-capturing amine’, *EPJ Quantum Technol.*, vol. 12, no. 1, p. 79, Dec. 2025, doi: 10.1140/epjqt/s40507-025-00385-8.

[55] C. Trahan, M. Loveland, and S. Dent, ‘Quantum physics-informed neural networks’, *Entropy*, vol. 26, no. 8, p. 649, Jul. 2024, doi: 10.3390/e26080649.

[56] Y. Xiao *et al.*, ‘Physics-informed quantum neural network for solving forward and inverse problems of partial differential equations’, *Physics of Fluids*, vol. 36, no. 9, p. 097145, Sep. 2024, doi: 10.1063/5.0226232.

[57] S. Berger, N. Hosters, and M. Möller, ‘Trainable embedding quantum physics informed neural networks for solving nonlinear PDEs’, *Sci Rep*, vol. 15, no. 1, p. 18823, May 2025, doi: 10.1038/s41598-025-02959-z.

[58] B. O. Fernandez, C. Leao, D. E. Revelo, and G. Fernandes, ‘Exploring hybrid physics-informed neural networks with quantum layers for solving the wave equation’, in *Fifth EAGE Digitalization Conference & Exhibition*, Edinburgh, Scotland, United Kingdom,: European Association of Geoscientists & Engineers, 2025, pp. 1–5. doi: 10.3997/2214-4609.202539060.

[59] N. B. Dehaghani, A. P. Aguiar, and R. Wisniewski, ‘A hybrid quantum-classical physics-informed neural network architecture for solving quantum optimal control problems’, 2024, *arXiv*. doi: 10.48550/ARXIV.2404.15015.

[60] F. Y. Leong, W.-B. Ewe, S. B. Q. Tran, Z. Zhang, and J. Y. Khoo, ‘Hybrid quantum physics-informed neural network: Towards efficient learning of high-speed flows’, *Computers & Fluids*, vol. 301, p. 106782, Oct. 2025, doi: 10.1016/j.compfluid.2025.106782.

[61] A. Farea, S. Khan, and M. S. Celebi, ‘QCPINN: Quantum-classical physics-informed neural networks for solving PDEs’, Oct. 18, 2025, *arXiv*: arXiv:2503.16678. doi: 10.48550/arXiv.2503.16678.

[62] Z. Song, R. Deaton, B. Gard, and S. H. Bryngelson, ‘Incompressible navier–stokes solve on noisy quantum hardware via a hybrid quantum–classical scheme’, *Computers & Fluids*, vol. 288, p. 106507, Feb. 2025, doi: 10.1016/j.compfluid.2024.106507.

[63] P. R. Hegde *et al.*, ‘A hybrid quantum-classical particle-in-cell method for plasma simulations’, *Future Generation Computer Systems*, vol. 175, p. 108087, Feb. 2026, doi: 10.1016/j.future.2025.108087.

[64] X. Rao, Y. Liu, and Y. Shen, ‘Quantum-classical physics-informed neural networks for solving reservoir seepage equations’, 2025, *arXiv*. doi: 10.48550/ARXIV.2512.03923.

[65] S. Lantigua, G. Giraldi, and R. Portugal, ‘Classical-quantum hybrid architecture for physics-informed neural networks’, *Phys. Rev. A*, vol. 113, no. 4, p. 042446, Apr. 2026, doi: 10.1103/fdd3-qz1s.

[66] Lloyd N. Trefethen, *Approximation Theory and Approximation Practice*. Philadelphia, Pennsylvania, USA: Society for Industrial and Applied Mathematics (SIAM), 2013.

[67] S. SS, K. AR, G. R, and A. KP, ‘Chebyshev polynomial-based kolmogorov-arnold networks: An efficient architecture for nonlinear function approximation’, Jun. 14, 2024, *arXiv*: arXiv:2405.07200. doi: 10.48550/arXiv.2405.07200.

[68] M. Larocca, N. Ju, D. García-Martín, P. J. Coles, and M. Cerezo, ‘Theory of overparametrization in

quantum neural networks', *Nat Comput Sci*, vol. 3, no. 6, pp. 542–551, Jun. 2023, doi: 10.1038/s43588-023-00467-6.

[69] Q. Wang, "Mathematical Foundations of Quantum Computing: Operator Theory, Entanglement, and Quantum Algorithms," diss., 2025. https://hdl.handle.net/1969.1/1598660.

[70] C. Canuto, M. Y. Hussaini, A. Quarteroni, and T. A. Zang, *Spectral methods: Evolution to complex geometries and applications to fluid dynamics*. in Scientific Computation Ser. Berlin, Heidelberg: Springer Berlin / Heidelberg, 2007.

[71] Z. Yu *et al.*, 'Non-asymptotic approximation error bounds of parameterized quantum circuits', in *Advances in Neural Information Processing Systems 37*, Vancouver, BC, Canada: Neural Information Processing Systems Foundation, Inc. (NeurIPS), 2024, pp. 99089–99127. doi: 10.52202/079017-3143.

[72] A. R. Barron, 'Universal approximation bounds for superpositions of a sigmoidal function', *IEEE Trans. Inform. Theory*, vol. 39, no. 3, pp. 930–945, May 1993, doi: 10.1109/18.256500.

[73] A. Jacot, F. Gabriel, and C. Hongler, 'Neural tangent kernel: Convergence and generalization in neural networks', 2018.

[74] J. Lazzari and X. Liu, 'Understanding the spectral bias of coordinate based MLPs via training dynamics', May 04, 2023, *arXiv*: arXiv:2301.05816. doi: 10.48550/arXiv.2301.05816.